\documentclass[acmsmall]{acmart}

\AtBeginDocument{%
  }

\usepackage{enumitem,xspace,multirow,makecell,threeparttable}
\usepackage[skins]{tcolorbox}
\usepackage{color}
\usepackage{colortbl}
\newcommand{\czp}[1]{{#1}}

\usepackage[framemethod=tikz]{mdframed}
\newcounter{finding}
\newcommand{\finding}[1]{\refstepcounter{finding}
  \vspace{2.3mm}
 \begin{mdframed}[linecolor=gray,roundcorner=12pt,backgroundcolor=gray!15,linewidth=3pt,innerleftmargin=2pt, leftmargin=0cm,rightmargin=0cm,topline=false,bottomline=false,rightline = false]
  \textbf{Finding \arabic{finding}:} #1
 \end{mdframed}
 \vspace{2.3mm}
}

\begin{document}

\title{Software Fairness Dilemma: Is Bias Mitigation a Zero-Sum Game?}

\author{Zhenpeng Chen}
\orcid{0000-0002-4765-1893}
\affiliation{%
  \institution{Nanyang Technological University}
  \city{Singapore}
  \country{Singapore}
}
\email{zhenpeng.chen@ntu.edu.sg}

\author{Xinyue Li}
\orcid{0009-0005-4058-6733}
\affiliation{%
  \institution{Peking University}
  \city{Beijing}
  \country{China}
}
\email{xinyueli@stu.pku.edu.cn}

\author{Jie M. Zhang}
\orcid{0000-0003-0481-7264}
\affiliation{%
  \institution{King's College London}
  \city{London}
  \country{United Kingdom}
}
\email{jie.zhang@kcl.ac.uk}

\author{Weisong Sun}\authornote{Corresponding author.}
\orcid{0000-0001-9236-8264}
\affiliation{%
  \institution{Nanyang Technological University}
  \city{Singapore}
  \country{Singapore}
}
\email{weisong.sun@ntu.edu.sg}

\author{Ying Xiao}
\orcid{0000-0002-8624-5740}
\affiliation{%
  \institution{King's College London}
  \city{London}
  \country{United Kingdom}
}
\email{ying.1.xiao@kcl.ac.uk}

\author{Tianlin Li}
\orcid{0000-0002-2207-1622}
\affiliation{%
  \institution{Nanyang Technological University}
  \city{Singapore}
  \country{Singapore}
}
\email{tianlin001@e.ntu.edu.sg}

\author{Yiling Lou}
\orcid{0000-0002-4066-3365}
\affiliation{%
  \institution{Fudan University}
  \city{Shanghai}
  \country{China}
}
\email{yilinglou@fudan.edu.cn}

\author{Yang Liu}
\orcid{0000-0001-7300-9215}
\affiliation{%
  \institution{Nanyang Technological University}
  \city{Singapore}
  \country{Singapore}
}
\email{yangliu@ntu.edu.sg}


\begin{abstract}
Fairness is a critical requirement for Machine Learning (ML) software, driving the development of numerous bias mitigation methods. Previous research has identified a leveling-down effect in bias mitigation for computer vision and natural language processing tasks, where fairness is achieved by lowering performance for all groups without benefiting the unprivileged group. However, it remains unclear whether this effect applies to bias mitigation for tabular data tasks, a key area in fairness research with significant real-world applications. This study evaluates eight bias mitigation methods for tabular data, including both widely used and cutting-edge approaches, across 44 tasks using five real-world datasets and four common ML models. Contrary to earlier findings, our results show that these methods operate in a zero-sum fashion, where improvements for unprivileged groups are related to reduced benefits for traditionally privileged groups. However, previous research indicates that the perception of a zero-sum trade-off might complicate the broader adoption of fairness policies. To explore alternatives, we investigate an approach that applies the state-of-the-art bias mitigation method solely to unprivileged groups, showing potential to enhance benefits of unprivileged groups without negatively affecting privileged groups or overall ML performance. Our study highlights potential pathways for achieving fairness improvements without zero-sum trade-offs, which could help advance the adoption of bias mitigation methods.
\end{abstract}



\maketitle

\section{Introduction}
Machine Learning (ML) software is increasingly adopted in critical decision-making domains, such as criminal sentencing, hiring, healthcare, and finance \cite{fairwaypaper,sigsoftChenZSH22,icseZhangH21}. However, it frequently exhibits discrimination based on sensitive attributes such as sex, race, and age \cite{ZPICSE24}. As a result, fairness has become a key requirement for ML software, drawing significant attention from software researchers and engineers \cite{jieMLsurvey,sigsoftBrunM18,reHorkoff19}. In response, the Software Engineering (SE) community has invested considerable efforts into developing bias mitigation methods \cite{ZPICSE24,fairwaypaper,fairsmotepaper,fairmaskpaper,xiao2024mirrorfair,icseLiMC0WZX22,softwareAlvarezM23,biswas2020machine,icseGaoZMSCW22}.

In the software fairness literature, ``unfairness'' typically measures performance disparities across demographic groups defined by sensitive attributes \cite{sigsoftChenZSH22,icseZhangH21,ZPICSE24}. Bias mitigation methods aim to reduce these disparities, thereby improving fairness.

However, this approach is often criticized as ``strictly egalitarian'' \cite{mittelstadt2023unfairness,emnlp4BDK23}, because it emphasizes relative performance between groups while neglecting absolute performance. This can lead to a phenomenon known as ``leveling down,'' where fairness is achieved by lowering performance for all groups without benefiting the unprivileged group \cite{wolff2001levelling,mittelstadt2023unfairness,emnlp4BDK23,ZietlowLBKLS022}. Zietlow et al. \cite{ZietlowLBKLS022} highlight this issue in bias mitigation methods for Computer Vision (CV) tasks, showing that these methods degrade performance for all groups, with the most significant drop for the privileged group. Similarly, Maheshwari et al. \cite{emnlp4BDK23} observe similar effects in CV and Natural Language Processing (NLP) tasks involving multiple sensitive attributes, where bias mitigation methods reduce the outcomes for privileged groups without benefiting unprivileged ones.

Despite these efforts, to the best of our knowledge, similar studies on tabular data tasks are lacking. Addressing this gap is important because bias mitigation methods for tabular data have critical applications in fairness-sensitive domains and represent the most extensively studied area in software fairness~\cite{DBcorrabs220707068}. \czp{Existing empirical studies on bias mitigation in tabular data tasks primarily focus on trade-offs between fairness and overall performance~\cite{sigsoftHortZSH21,Dabs220703277,biswas2020machine}. In contrast, this paper aims to investigate trade-offs in group benefits.}

We conduct a large-scale empirical study on eight representative bias mitigation methods for tabular data, including both widely used and cutting-edge techniques. We evaluate these methods across 44 tabular data tasks spanning social, financial, and medical domains, using five real-world datasets and four common ML models, while considering scenarios with both single and multiple sensitive attributes.

Unlike previous findings in CV and NLP \cite{ZietlowLBKLS022,emnlp4BDK23}, our study reveals that bias mitigation methods for tabular data exhibit a zero-sum trade-off, which indicates that gains for one group result in corresponding losses for another \cite{norton2011whites,brown2022if}. Specifically, we observe that improvements for unprivileged groups are accompanied by reductions in benefits for traditionally privileged groups. For example, existing bias mitigation methods significantly increase the selection rate (i.e., the rate of assigning favorable outcomes) and true positive rate for unprivileged groups by large effect sizes in 37.5\% to 84.4\% and 21.9\% to 87.5\% of tasks, respectively. Meanwhile, these methods cause significant reductions in the selection rate and true positive rate for privileged groups, with large effect sizes in 31.3\% to 62.5\% and 25.0\% to 62.5\% of tasks, respectively. Furthermore, greater fairness improvements are significantly associated with larger gains in the selection and true positive rates for unprivileged groups, but also with more substantial decreases in these rates for privileged groups.

However, previous research \cite{brown2022if, norton2011whites} suggests that the perception of a zero-sum trade-off may hinder the broader adoption of fairness policies. This belief can pose a significant challenge, as even individuals with strong egalitarian values may resist fairness initiatives if they believe that improving outcomes for one group comes at the expense of another \cite{brown2022if}. As a result, the zero-sum trade-offs observed in bias mitigation methods for tabular data not only present technical challenges but may also exacerbate sociopolitical barriers, potentially limiting the widespread acceptance of these methods.

This motivates us to explore whether it is possible to avoid a zero-sum trade-off in bias mitigation methods for tabular data, thereby facilitating the adoption of these methods in practice. To this end, we investigate applying the state-of-the-art bias mitigation method solely to unprivileged groups. This approach enhances benefits for unprivileged groups while preserving the benefits for privileged groups and maintaining overall ML performance (e.g., accuracy and F1-score) comparable to applying the method universally. Additionally, the approach results in a marginal increase in the overall selection rate, averaging just 0.01, compared to applying the method across the entire population. This study serves as a preliminary exploration of pathways toward achieving fairness improvements without incurring zero-sum trade-offs.

In summary, this paper makes the following contributions:

\begin{itemize}[leftmargin=*]
\item We conduct a large-scale empirical study to characterize the impact of existing bias mitigation methods for tabular data on different demographic groups.
\item We perform a preliminary investigation into the possibility of avoiding the zero-sum trade-off in bias mitigation for tabular data.
\item We provide a replication package \cite{githublink} that includes all data and code used in the paper to support replication and further research.
\end{itemize}

\section{Preliminaries}\label{prelimi}
We begin by providing background knowledge on ML software fairness \czp{and discussing the relevance of bias mitigation to SE}, followed by a review of related work.

\subsection{ML Software Fairness}\label{prefairness}
We focus on ML classification for tabular data, the most widely studied area in software fairness research~\cite{icseZhangH21,sigsoftChenZSH22,sigsoftHortZSH21,ZPICSE24,fairwaypaper,fairsmotepaper,fairmaskpaper,sigsoftBiswasR21,biswas2020machine,xiao2024mirrorfair,icseNiariKT022}, particularly because of its crucial role in fields such as finance, healthcare, and criminal justice, where fairness is essential. Decision-making involving tabular data often includes explicit sensitive attributes such as race, sex, and age, which, if mishandled, can lead to discriminatory outcomes.

Sensitive attributes can create divisions between privileged and unprivileged groups. In practice, ML software often favors privileged groups by assigning them positive labels, while unprivileged groups receive unfavorable labels. For example, an income prediction system might be more likely to predict high income for males (privileged group) and low income for females (unprivileged group), with sex as the sensitive attribute, high income as the favorable label, and low income as the unfavorable label.

Such bias has driven extensive research into bias mitigation to improve the group fairness of ML software~\cite{sigsoftChenZSH22,fairwaypaper,fairsmotepaper,fairmaskpaper,xiao2024mirrorfair}, which is also the focus of our paper. Group fairness, widely encoded in laws and regulations, is recognized as a crucial non-functional requirement of ML software \cite{Dabs220710223}. It ensures that different demographic groups, defined by sex, race, age, or other sensitive attributes, are treated equitably by ML software, preventing biased decisions that favor one group over others.

To quantitatively measure group fairness, various metrics have been proposed, including Statistical Parity Difference (SPD), Equal Opportunity Difference (EOD), and Average Odds Difference (AOD), all of which are widely adopted in both literature and practice~\cite{DBcorrabs220707068}.
These metrics share a common motivation: to balance classification performance across different demographic groups.

Let $A$ denote a sensitive attribute, where 1 represents the privileged group and 0 the unprivileged group. $Y$ represents the actual label, and $\hat{Y}$ the predicted label, with 1 indicating the favorable label and 0 the unfavorable label. Below are the definitions and calculations of these fairness metrics:

SPD measures the difference in selection rates between the privileged and unprivileged groups.

\begin{equation}
\footnotesize
\begin{aligned}
SPD = \vert P[\hat{Y} = 1 | A = 1] - P[\hat{Y} = 1 | A = 0] \vert.
\end{aligned}
\end{equation}

EOD measures the difference in true positive rates between the privileged and unprivileged groups.

\begin{equation}
\footnotesize
\begin{aligned}
EOD = \vert P[\hat{Y}=1| A=1, Y=1]  -  P[\hat{Y}=1|A=0, Y=1] \vert.
\end{aligned}
\end{equation}

AOD measures the average of the differences in true positive rates and false positive rates between the privileged and unprivileged groups.

\begin{equation}
\footnotesize
\begin{aligned}
AOD  = \vert\frac{1}{2}[(P[\hat{Y}=1| A=1, Y=1]  -  P[\hat{Y}=1|A=0, Y=1]) +
(P[\hat{Y}=1| A=1, Y=0]  -  P[\hat{Y}=1|A=0, Y=0])]\vert.
\end{aligned}
\end{equation}

To extend these fairness metrics to scenarios with multiple sensitive attributes, we divide the population into multiple demographic groups based on the sensitive attributes considered. For example, with two sensitive attributes, i.e., sex (male, female) and race (white, non-white), we generate four groups: white male, white female, non-white male, and non-white female. We then quantify the maximum disparity between these groups by measuring the difference among the groups with the least and greatest discrimination \cite{ZPICSE24}. Specifically, for SPD, we calculate the maximum difference in selection rates among the groups; for EOD, the maximum difference in true positive rates; and for AOD, the maximum average difference in true positive rates and false positive rates. Due to the page limit, we omit the detailed equations.

\czp{\subsection{Relevance of Bias Mitigation to SE}
Bias mitigation is a highly relevant topic to SE, holding significant importance for software researchers, engineers, and companies. \textbf{(1) For software researchers:} Fairness is a critical non-functional software requirement~\cite{sigsoftBrunM18,reHorkoff19,jieMLsurvey}, and unfairness in ML software is regarded as a fairness bug by the SE community~\cite{Dabs220710223,sigsoftChenZSH22}. Thus, bias mitigation (i.e., addressing fairness bugs) has received substantial attention from SE researchers. Five of the bias mitigation methods we study (Section~\ref{methodsection}) were introduced in top SE venues (ICSE, FSE, and TSE). Notably, FairSMOTE~\cite{fairsmotepaper}, one of these methods, earned the Distinguished Paper Award at FSE’21, underscoring its impact and relevance to SE. \textbf{(2) For software engineers:} Ensuring fairness in ML software is widely recognized as an ethical duty for software engineers~\cite{fairsmotepaper,ZPICSE24}, particularly as they integrate ML models into software systems while adhering to ethical and legal standards. \textbf{(3) For software companies:} Unfair ML software poses significant risks to software companies, including ethical, reputational, financial, and legal consequences, particularly if they violate anti-discrimination laws~\cite{Dabs220703277,icseZhangH21}.
}

\subsection{Related Work}
\noindent \textbf{Bias Mitigation Methods.} Numerous bias mitigation methods have been proposed to enhance group fairness. These methods are generally categorized into three types: pre-processing, in-processing, and post-processing \cite{DBcorrabs220707068}.
(1) \emph{Pre-processing methods} reduce bias in the training data, thereby improving the fairness of the resulting ML models. For instance, Chakraborty et al. \cite{fairsmotepaper} identified bias in prior decisions as a root cause of ML software bias, and addressed this by rebalancing internal data distributions and removing biased labels from the training data.
(2) \emph{In-processing methods} enhance fairness during the model training process. For example, Gao et al. \cite{icseGaoZMSCW22} proposed a method specifically for deep neural networks, which involves detecting neurons that exhibit contradictory optimization directions for accuracy and fairness, followed by selective dropout training to mitigate bias.
(3) \emph{Post-processing methods} adjust the model's outcomes to ensure fairer results. For example, Peng et al. \cite{fairmaskpaper} modified the values of sensitive attributes in the test data to make the model's decisions fair.
\czp{There are also hybrid approaches that span different bias mitigation types. For example, Xiao et al.~\cite{xiao2024mirrorfair} proposed MirrorFair, which first processes the training dataset to generate a counterfactual dataset, then trains models separately on both datasets, and finally adaptively combines their predictions to adjust the model’s outcomes. Works like MirrorFair~\cite{xiao2024mirrorfair} present technical contributions by proposing new bias mitigation methods. In contrast, our paper makes an empirical contribution by systematically examining the zero-sum trade-offs between groups during bias mitigation, an aspect not addressed in these papers.} 

\noindent \textbf{Empirical Evaluation.} To systematically understand the effects of bias mitigation methods, researchers have conducted numerous empirical investigations. Chen et al. \cite{ZPICSE24} identified a side effect of bias mitigation: addressing bias concerning one sensitive attribute can inadvertently amplify bias regarding others. Biswas and Rajan \cite{biswas2020machine} applied bias mitigation techniques to ML models sourced from Kaggle, a crowd-sourced platform, to analyze their impact on fairness. Recognizing that fairness improvements often come at the cost of ML performance, they also assessed the impact of these methods on ML performance. Hort et al. \cite{sigsoftHortZSH21} and Chen et al. \cite{Dabs220703277} conducted studies to characterize the fairness-performance trade-off achieved by existing bias mitigation methods. Similarly, Friedler et al.~\cite{FriedlerSVCHR19} and Menon and Williamson~\cite{fatMenonW18} studied the trade-offs between fairness and overall accuracy (an important ML performance metric).

All these studies focus on bias mitigation for tabular data, which is the most extensively studied in the software fairness literature \cite{DBcorrabs220707068}. In contrast, Yang et al. \cite{zeyujjj} conducted an empirical study of bias mitigation methods specifically for image classification tasks, evaluating their effects on both fairness and ML performance.

These empirical studies primarily focus on overall performance without providing a more in-depth analysis of how performance varies across different demographic groups. To address this gap, a few studies have been conducted. Zietlow et al. \cite{ZietlowLBKLS022} evaluated bias mitigation methods in CV tasks and found that these methods improve fairness by degrading performance across all groups. Similarly, Maheshwari et al. \cite{emnlp4BDK23} conducted an empirical study on bias mitigation methods for CV and NLP tasks involving multiple sensitive attributes, noting that these methods improve fairness by harming the best-off group without benefiting the worst-off group. 
In contrast, there is a lack of similar empirical studies on tabular data, despite its prominence in fairness research. 
Zafar et al.~\cite{ZafarVGG17} introduced a fairness measure related to disparate mistreatment and explored its implications for group benefits, but they did not evaluate the trade-offs in group benefits achieved by existing bias mitigation methods.

\section{Experimental Setup}
This section describes our research questions and experimental setup.

\subsection{Research Questions (RQs)}
We aim to answer the following RQs in this study.

\noindent \textbf{RQ1:} \emph{How do bias mitigation methods impact various demographic groups in tabular data tasks?} Existing studies have addressed this question on CV and NLP tasks \cite{ZietlowLBKLS022,emnlp4BDK23}, but this paper shifts the focus to tabular data, which is the most extensively studied topic in software fairness research~\cite{DBcorrabs220707068}.

\noindent \textbf{RQ2:} \emph{What is the correlation between the impact on different demographic groups and the overall impact on fairness?} Since bias mitigation methods aim to enhance fairness, it is important to quantitatively assess how changes in fairness correlate with the effects on different demographic groups. This analysis can provide deeper insights into the relationship between fairness improvements and group-specific outcomes.

\noindent \textbf{RQ3:} \emph{How do bias mitigation methods impact demographic groups defined by multiple sensitive attributes?} As research increasingly considers multiple sensitive attributes simultaneously \cite{ZPICSE24}, the complexity of assessing bias mitigation methods grows. This RQ explores how bias mitigation methods for tabular data perform on each group when multiple attributes are considered, leading to a more nuanced understanding of their effects across intersecting demographic groups.

\noindent \textbf{RQ4:} \emph{Is it possible to improve fairness without degrading the performance of either privileged or unprivileged groups?} Bias mitigation methods that do not lead to negative outcomes for any group are likely to be more widely adopted by society. This question investigates whether such an ideal situation is achievable and explores alternative approaches that might allow for fairness improvements without detriment to any group.

\subsection{Bias Mitigation Methods}\label{methodsection}
We use a total of eight existing bias mitigation methods for our study. On one hand, we employ three most widely adopted bias mitigation methods, as identified in a recent survey~\cite{DBcorrabs220707068}: Adversarial Debiasing (ADV)~\cite{ADVpaper}, Reweighting (REW)~\cite{rewpaper}, and Equalized Odds Processing (EOP)~\cite{EOpaper}. On the other hand, we include five recently proposed methods from the software fairness literature: FairSMOTE~\cite{fairsmotepaper}, LTDD~\cite{icseLiMC0WZX22}, MAAT~\cite{sigsoftChenZSH22}, FairMask~\cite{fairmaskpaper}, and MirrorFair~\cite{xiao2024mirrorfair}. 
\czp{Our selection spans pre-processing, in-processing, and post-processing approaches, ensuring a comprehensive evaluation.}

In the following, we provide a brief introduction to each method.

\begin{itemize}[leftmargin=*]
\item ADV~\cite{ADVpaper} uses adversarial learning to make predictions less dependent on sensitive attributes during the training process.
\item REW~\cite{rewpaper} adjusts the weights of different instances in the training data to ensure that underrepresented groups are given more importance.
\item EOP~\cite{EOpaper} adjusts model predictions to ensure that the false positive and false negative rates are similar across different demographic groups.
\item FairSMOTE~\cite{fairsmotepaper} balances internal distributions of training data concerning both sensitive attributes and class labels, while also removing biased labels.
\item LTDD~\cite{icseLiMC0WZX22} is a data debugging method that identifies and excludes biased parts of features in the training data for building fair ML software.
\item MAAT~\cite{sigsoftChenZSH22} trains individual models optimized separately for ML performance and fairness, subsequently combining their predictions for a balanced fairness-performance trade-off.
\item FairMask~\cite{fairmaskpaper} trains extrapolation models to predict sensitive attributes based on other features and uses these models to relabel sensitive attributes during the inference process.
\item MirrorFair~\cite{xiao2024mirrorfair} constructs a counterfactual dataset from the original data, trains models on both datasets, and adaptively combines their predictions for fair decisions.
\end{itemize}

\subsection{Bias Mitigation Tasks}\label{taskslist}
We use the same set of 44 bias mitigation tasks as previous work \cite{xiao2024mirrorfair}, which are generated using five real-world datasets and four widely used ML models.

\noindent \textbf{Datasets.}
Table \ref{dataset_info} presents an overview of the datasets used in our study.
\czp{Specifically, we use five real-world tabular datasets extensively employed in fairness research~\cite{sigsoftChenZSH22,Dabs220703277,xiao2024mirrorfair,icseZhangH21}, spanning different domains and fairness-critical scenarios: income prediction, recidivism prediction, credit prediction, deposit subscription prediction, and healthcare needs prediction.}
These datasets span the financial, social, and medical domains, providing a broad scope for analysis. 

For each dataset, we use the sensitive attributes as defined within it. These datasets cover sex, race, and age, which are recognized as the three most commonly considered sensitive attributes~\cite{DBcorrabs220707068}.

\czp{We follow previous work~\cite{xiao2024mirrorfair} to transform the five datasets into 11 tasks: 8 single-attribute tasks (Adult-Sex, Adult-Race, Compas-Sex, Compas-Race, German-Sex, German-Age, Bank-Age, and Mep-Race) and 3 multiple-attribute tasks (Adult-Sex-Race, Compas-Sex-Race, and German-Sex-Age).}

\noindent \textbf{ML Models.} For each task, we train four representative ML models: Logistic Regression (LR), Random Forest (RF), Support Vector Machine (SVM), and Deep Neural Network (DNN). \czp{These models are demonstrated to be the most extensively studied in fairness research~\cite{DBcorrabs220707068} and widely used in fairness-critical decision applications \cite{ZPICSE24}.} For LR, RF, and SVM, we use the configurations specified in recent studies~\cite{ZPICSE24,Dabs220703277,sigsoftChenZSH22,sigsoftHortZSH21}. For DNN, we employ a fully connected network with five hidden layers comprising 64, 32, 16, 8, and 4 units, respectively, a model architecture extensively used for the datasets we study~\cite{ZPICSE24,Dabs220703277,sigsoftZhang022, icseZhengCD0CJW0C22}.

\czp{By converting the datasets into eight single-attribute tasks and three multiple-attribute tasks, and training four models for each, we finally obtain $8\times4=32$ single-attribute bias mitigation tasks and $3\times4=12$ multiple-attribute bias mitigation tasks.}

\begin{table}[!tp]
\scriptsize
\centering
\caption{Datasets.}
\label{dataset_info}
\begin{tabular}{lrlll}
\hline
Name & Size & Sensitive attr(s) & Favorable label & Description\\
\hline
Adult  & 45,222  & sex, race & income $\textgreater$ 50K & predict whether an individual’s income exceeds 50K \\
Compas & 6,167 & sex, race & no recidivism & predict recidivism for criminal defendants\\
German & 1,000 & sex, age & good credit & predict whether an individual has
good credit \\
Bank  & 30,488 & age & subscriber & predict whether an individual will subscribe to a deposit \\
Mep & 15,830 & race & utilizer & predict individual healthcare needs \\
\hline
\end{tabular}
\end{table}

\subsection{Evaluation Metrics}\label{emetric}
We use the same set of fairness metrics as in previous work \cite{sigsoftChenZSH22,ZPICSE24}, including SPD, EOD, and AOD.
\czp{The selection of fairness metrics aligns with established practices in the fairness literature, covering the most widely adopted metrics in the field according to a recent survey~\cite{DBcorrabs220707068}.} Detailed descriptions of these metrics are provided in Section~\ref{prelimi}.

\czp{To explore the zero-sum trade-offs between the benefits for privileged and unprivileged groups, we disaggregate the fairness metrics by focusing on three group-level performance metrics: selection rate (SR), true positive rate (TPR), and false positive rate (FPR). These performance metrics are used by the fairness metrics to measure the benefits for each group.}
Higher SR and TPR values indicate better performance and greater benefits, while a lower FPR value typically suggests better performance. However, it is important to note that a high FPR for a demographic group does not mean the group is being harmed. In bias mitigation tasks, the positive label represents a favorable outcome, so a high FPR indicates that the bias mitigation method tends to favor this group, even assigning favorable outcomes to the group members who are not qualified.

For single-attribute tasks, we denote these rates for the privileged group as $SR_P$, $TPR_P$, and $FPR_P$, and for the unprivileged group as $SR_U$, $TPR_U$, and $FPR_U$. In multiple-attribute tasks, demographic groups are ranked based on the proportion of members achieving favorable outcomes in the training data, producing groups labeled $Group_1$, $Group_2$, ..., $Group_n$, where $Group_1$ is the most favored and $Group_n$ is the least favored. The performance for $Group_i$ is denoted by $SR_i$, $TPR_i$, and $FPR_i$.

\subsection{Implementation Details}
To ensure the reproducibility of our results, we outline the implementation details. Consistent with previous work \cite{ZPICSE24}, each bias mitigation method is applied to each task 20 times, with performance and fairness metrics averaged across these runs to mitigate the impact of randomness. In each experiment, the dataset is randomly split, with 70\% used as training data and 30\% as test data, following a common practice in the software fairness literature \cite{ZPICSE24,sigsoftChenZSH22,xiao2024mirrorfair}. 

\section{RQ1: Impact on Different Demographic Groups}\label{rq1}
In this RQ, we apply bias mitigation methods to 32 single-attribute bias mitigation tasks. In such tasks, the population is typically divided into two groups: a privileged group and an unprivileged group. We examine the SR, TPR, and FPR of the two groups, both with and without the application of bias mitigation methods, to assess their impact on the performance for different groups. Specifically, we analyze the frequency of various impact types (e.g., performance decrease or increase) across the 32 tasks (\textbf{RQ1.1}), followed by an analysis of the effect size of these impacts (\textbf{RQ1.2}).

\subsection{RQ1.1: Frequency Analysis}
\noindent \textbf{Methodology.}
To determine whether the performance for a group is significantly impacted by a bias mitigation method, we use the non-parametric Mann-Whitney U-test \cite{mcknight2010mann}, a widely adopted approach in software fairness research \cite{Dabs220703277, xiao2024mirrorfair, ZPICSE24}. Following established practices \cite{xiao2024mirrorfair, ZPICSE24, Dabs220703277}, we consider an impact to be statistically significant only if the $p$-value from the test is below 0.05, ensuring our findings have a 95\% confidence level. For example, when comparing two sets of SR values for the unprivileged group across 20 runs, with and without the bias mitigation method, the null hypothesis assumes no significant difference in SR between the two conditions. If the test yields a $p$-value < 0.05, we can reject the null hypothesis and conclude with 95\% confidence that applying the bias mitigation method significantly affects the SR of the unprivileged group.

We classify impact types as follows: an increase in the average metric value with a $p$-value < 0.05 is considered a significant increase; a decrease with a $p$-value < 0.05 is considered a significant decrease; and if the $p$-value $\geq$ 0.05, the result is considered a tie. For each bias mitigation method, we calculate the number of scenarios in which it leads to a significant increase, tie, or significant decrease in each group performance metric.

\noindent \textbf{Results.} Table \ref{frequencytable1} presents the results. Overall, bias mitigation methods tend to decrease SR, TPR, and FPR for the privileged group while increasing them for the unprivileged group, thereby narrowing the performance gap and improving fairness. Initially, the original models show higher SR, TPR, and FPR for the privileged group compared to the unprivileged group. 
The bias mitigation methods we study significantly reduce $SR_P$, $TPR_P$, and $FPR_P$ in 52.0\% (133/256), 50.8\% (130/256), and 45.7\% (117/256) of tasks, respectively, while significantly increasing them in 7.4\% (19/256), 7.4\% (19/256), and 6.6\% (17/256) of tasks. In contrast, these methods significantly increase $SR_U$, $TPR_U$, and $FPR_U$ in 64.1\% (164/256), 55.5\% (142/256), and 60.9\% (156/256) of tasks, while significantly decreasing them in 3.1\% (8/256), 4.3\% (11/256), and 2.7\% (7/256) of tasks.

This pattern holds across all eight methods studied. Each method more frequently decreases $SR_P$, $TPR_P$, and $FPR_P$ than increases them, while it more frequently increases $SR_U$, $TPR_U$, and $FPR_U$ than decreases them. For instance, the state-of-the-art MirrorFair decreases $SR_P$ in 46.9\% (15/32) of tasks but increases it in only 6.3\% (2/32). Conversely, these methods are more likely to increase $SR_U$, $TPR_U$, and $FPR_U$. For example, MirrorFair increases $SR_U$ in 87.5\% (28/32) of tasks, with a decrease in just 6.3\% (2/32).

In summary, we present two key observations.
\begin{itemize}[leftmargin=*]
\item Unlike previous bias mitigation studies in CV and NLP, where fairness improvements typically result from reducing performance for both groups (with a larger reduction for the privileged group), we do not observe the same leveling-down effect in tabular data tasks. Instead, bias mitigation methods for tabular data exhibit a zero-sum trade-off, where they improve SR, TPR, and FPR for the unprivileged group while reducing them for the privileged group. This zero-sum pattern is consistent across all eight methods and all three metrics (SR, TPR, and FPR).
\item The choice of performance metric is critical. Since higher SR and TPR, and lower FPR, generally indicate better performance, we conclude that current bias mitigation methods for tabular data improve SR and TPR for the unprivileged group at the expense of the privileged group, but decrease the unprivileged group's performance in terms of FPR. However, as discussed in Section~\ref{emetric}, higher FPR can also indicate greater benefits in bias mitigation contexts. Thus, the overall effect of these methods is to promote fairness by enhancing benefits for the unprivileged group, even if this comes at a cost to the privileged group.
\end{itemize}

\finding{Contrary to previous findings in CV and NLP, where bias mitigation methods typically improve fairness by reducing performance for both groups (a phenomenon known as leveling down), our study on tabular data tasks reveals a zero-sum trade-off. Specifically, we find that bias mitigation methods tend to adjust the selection rate, true positive rate, and false positive rate by lowering these metrics for the privileged group and raising them for the unprivileged group, thus narrowing disparities and enhancing fairness. Notably, this pattern holds consistently across all eight methods we examine.}

\begin{table*}[!tp]
\scriptsize
\centering
\caption{(RQ1.1) Number of tasks where each bias mitigation method significantly increases ($\uparrow$), decreases ($\downarrow$), or does not significantly impact ($-$) the SR, TPR, and FPR for each group. Overall, bias mitigation methods for tabular data tend to decrease SR, TPR, and FPR for the privileged group while increasing them for the unprivileged group.}
\label{frequencytable1}
\begin{tabular}{l|rrr|rrr}
\hline
Method & $SR_P$ ($\uparrow$) & $SR_P$ ($-$) & $SR_P$ ($\downarrow$) & $SR_U$ ($\uparrow$) & $SR_U$ ($-$) & $SR_U$ ($\downarrow$) \\
\hline
ADV & 4 & 14 & 14 & 16 & 12 & 4 \\
REW & 0 & 10 & 22 & 21 & 11 & 0\\
EOP & 1 & 14 & 17 & 22 & 10 & 0\\
FairSMOTE & 8 & 4 & 20 & 24 & 8 & 0\\
LTDD & 2 & 15 & 15 & 17 & 13 & 2\\
MAAT & 2 & 11 & 19 & 21 & 11 & 0\\
FairMask & 0 & 21 & 11 & 15 & 17 & 0\\
MirrorFair & 2 & 15 & 15 & 28 & 2 & 2\\
\hline
\rowcolor{gray!15}
Overall & 19 & 104 & 133 & 164 & 84 & 8\\
\hline
\hline
Method & $TPR_P$ ($\uparrow$) & $TPR_P$ ($-$) & $TPR_P$ ($\downarrow$) & $TPR_U$ ($\uparrow$) & $TPR_U$ ($-$) & $TPR_U$ ($\downarrow$)\\
\hline
ADV & 4 & 11 & 17& 16 & 12 & 4 \\
REW & 0 & 12 & 20 & 18 & 14 & 0\\
EOP & 1 & 12 & 19 & 14 & 15 & 3\\
FairSMOTE & 9 & 2 & 21 & 21 & 11 & 0\\
LTDD & 1 & 17 & 14 & 16 & 14 & 2\\
MAAT & 2 & 14 & 16 & 19 & 13 & 0\\
FairMask & 0 & 22 & 10 & 10 & 22 & 0\\
MirrorFair & 2 & 17 & 13 & 28 & 2 & 2 \\
\hline
\rowcolor{gray!15}
Overall & 19 & 107 & 130 & 142 & 103 & 11\\
\hline
\hline
Method & $FPR_P$ ($\uparrow$) & $FPR_P$ ($-$) & $FPR_P$ ($\downarrow$) & $FPR_U$ ($\uparrow$) & $FPR_U$ ($-$) & $FPR_U$ ($\downarrow$)\\
\hline
ADV & 3 & 18 & 11 & 15 & 13 & 4 \\
REW & 0 & 12 & 20 & 20 & 12 & 0 \\
EOP & 2 & 19 & 11 & 22 & 10 & 0 \\
FairSMOTE & 8 & 6 & 18 & 23 & 9 & 0\\
LTDD & 2 & 16 & 14 & 17 & 14 & 1\\
MAAT & 2 & 12 & 18 & 20 & 12 & 0\\
FairMask & 0 & 21 & 11 & 12 & 20 & 0\\
MirrorFair & 0 & 18 & 14 & 27 & 3 & 2 \\
\hline
\rowcolor{gray!15}
Overall & 17 & 122 & 117 & 156 & 93 & 7\\
\hline
\end{tabular}
\end{table*}

\subsection{RQ1.2: Effect Size Analysis}
\noindent \textbf{Methodology.} To further analyze the impact of bias mitigation methods on each group, we employ Cliff's $\delta$ \cite{eseKitchenhamMBKBC17}, a widely used effect size metric in SE research \cite{eseKitchenhamMBKBC17, ZPICSE24, esemBenninKMPM17}. Following standard practice~\cite{eseKitchenhamMBKBC17, ZPICSE24, esemBenninKMPM17}, we interpret a $\delta$ value with an absolute magnitude of 0.428 or greater as indicating a large effect size. First, we determine the proportion of tasks where each method significantly decreases SR, TPR, and FPR by a large effect size. Second, for each method, we calculate the mean and maximum decrease in $SR_P$, $TPR_P$, and $FPR_P$ across 32 single-attribute tasks. We perform a similar analysis to assess the effect size of decreases for $SR_U$, $TPR_U$, and $FPR_U$.

\begin{table*}[!tp]
\tiny
\centering
\caption{(RQ1.2) Effect size of bias mitigation methods on performance for each group. The table presents the mean impact and the maximum performance decrease/increase for the privileged/unprivileged groups, \czp{along with how these relative changes are derived from absolute numbers. For example, the mean impact of ADV in $SR_P$ is -0.026 (0.462-0.489), indicating that ADV reduces the average $SR_P$ across tasks from 0.489 to 0.462.} The table also shows the proportions of tasks where each method results in a significantly large performance decrease/increase for these groups. We find that existing bias mitigation methods substantially increase the benefits for the unprivileged group, with significant increases in $SR_U$ and $TPR_U$ in 37.5\%$\sim$84.4\% and 21.9\%$\sim$87.5\% of tasks, respectively. However, these methods also lead to significant reductions in the $SR_P$ and $TPR_P$, with large effect sizes observed in 31.3\%$\sim$62.5\% and 25.0\%$\sim$62.5\% of tasks, respectively.}
\label{effectsizetable}
\begin{tabular}{l|rrr|rrr|rrr}
\hline
\multirow{2}*{Method}&\multicolumn{3}{c|}{$SR_P$}&\multicolumn{3}{c|}{$TPR_P$}&\multicolumn{3}{c}{$FPR_P$}\\
& Mean & Max $\downarrow$ & Large $\downarrow$ & Mean & Max $\downarrow$ & Large $\downarrow$ & Mean & Max $\downarrow$ & Large $\downarrow$ \\
\hline
ADV & \makecell[r]{-0.026\\\czp{(0.462-0.489)}} & \makecell[r]{-0.126\\\czp{(0.716-0.841)}} & 43.8\% & \makecell[r]{-0.049\\\czp{(0.654-0.704)}} & \makecell[r]{-0.175\\\czp{(0.460-0.635)}} & 50.0\% & \makecell[r]{0.000\\\czp{(0.334-0.334)}} & \makecell[r]{-0.085\\\czp{(0.497-0.582)}} & 34.4\%\\
REW & \makecell[r]{-0.043\\\czp{(0.446-0.489)}} & \makecell[r]{-0.177\\\czp{(0.651-0.828)}} & 59.4\% & \makecell[r]{-0.058\\\czp{(0.646-0.704)}} & \makecell[r]{-0.189\\\czp{(0.350-0.539)}} & 56.3\% & \makecell[r]{-0.043\\\czp{(0.291-0.334)}} & \makecell[r]{-0.213\\\czp{(0.476-0.690)}} & 56.3\%\\
EOP & \makecell[r]{-0.024\\\czp{(0.465-0.489)}} & \makecell[r]{-0.110\\\czp{(0.698-0.808)}} & 37.5\% & \makecell[r]{-0.043\\\czp{(0.660-0.704)}} & \makecell[r]{-0.162\\\czp{(0.320-0.482)}} & 46.9\% &  \makecell[r]{-0.011\\\czp{(0.323-0.334)}} & \makecell[r]{-0.075\\\czp{(0.493-0.568)}} & 34.4\%\\
FairSMOTE & \makecell[r]{-0.026\\\czp{(0.462-0.489)}} & \makecell[r]{-0.156\\\czp{(0.672-0.828)}} & 62.5\% & \makecell[r]{-0.035\\\czp{(0.669-0.704)}} & \makecell[r]{-0.357\\\czp{(0.182-0.539)}} & 62.5\% & \makecell[r]{-0.035\\\czp{(0.299-0.334)}} & \makecell[r]{-0.180\\\czp{(0.392-0.572)}} & 56.3\%\\
LTDD & \makecell[r]{-0.033\\\czp{(0.456-0.489)}} & \makecell[r]{-0.212\\\czp{(0.616-0.828)}} & 40.6\% & \makecell[r]{-0.044\\\czp{(0.659-0.704)}} & \makecell[r]{-0.219\\\czp{(0.320-0.539)}} & 40.6\% & \makecell[r]{-0.032\\\czp{(0.303-0.334)}} & \makecell[r]{-0.245\\\czp{(0.445-0.690)}} & 37.5\%\\
MAAT & \makecell[r]{-0.025\\\czp{(0.464-0.489)}} & \makecell[r]{-0.078\\\czp{(0.196-0.273)}} & 56.3\% & \makecell[r]{-0.043\\\czp{(0.660-0.704)}} & \makecell[r]{-0.138\\\czp{(0.477-0.615)}} & 50.0\% & \makecell[r]{-0.019\\\czp{(0.315-0.334)}} & \makecell[r]{-0.051\\\czp{(0.058-0.109)}} & 53.1\%\\
FairMask & \makecell[r]{-0.015\\\czp{(0.474-0.489)}} & \makecell[r]{-0.062\\\czp{(0.766-0.828)}} & 31.3\% & \makecell[r]{-0.014\\\czp{(0.689-0.704)}} & \makecell[r]{-0.056\\\czp{(0.722-0.778)}} & 25.0\% & \makecell[r]{-0.020\\\czp{(0.314-0.334)}} & \makecell[r]{-0.093\\\czp{(0.597-0.690)}} & 21.9\%\\
MirrorFair & \makecell[r]{-0.015\\\czp{(0.474-0.489)}} & \makecell[r]{-0.063\\\czp{(0.764-0.828)}} & 46.9\% & \makecell[r]{-0.018\\\czp{(0.685-0.704)}} & \makecell[r]{-0.098\\\czp{(0.517-0.615)}} & 37.5\% & \makecell[r]{-0.017\\\czp{(0.317-0.334)}} & \makecell[r]{-0.080\\\czp{(0.610-0.690)}} & 43.8\%\\
\hline
\hline
\multirow{2}*{Method}&\multicolumn{3}{c|}{$SR_U$}&\multicolumn{3}{c|}{$TPR_U$}&\multicolumn{3}{c}{$FPR_U$}\\
& Mean & Max $\uparrow$ & Large $\uparrow$ & Mean & Max $\uparrow$ & Large $\uparrow$ & Mean & Max $\uparrow$ & Large $\uparrow$ \\
\hline
ADV	& \makecell[r]{0.006\\\czp{(0.354-0.348)}} & \makecell[r]{0.081\\\czp{(0.198-0.116)}} & 50.0\% & \makecell[r]{0.043\\\czp{(0.649-0.605)}} & \makecell[r]{0.226\\\czp{(0.719-0.493)}} & 46.9\% & \makecell[r]{0.006\\\czp{(0.235-0.229)}} & \makecell[r]{0.057\\\czp{(0.096-0.039)}} & 46.9\%\\
REW& \makecell[r]{0.036\\\czp{(0.385-0.348)}} & \makecell[r]{0.144\\\czp{(0.750-0.606)}} & 62.5\% & \makecell[r]{0.055\\\czp{(0.661-0.605)}} & \makecell[r]{0.187\\\czp{(0.680-0.493)}} & 56.3\% & \makecell[r]{0.038\\\czp{(0.267-0.229)}} & \makecell[r]{0.180\\\czp{(0.613-0.433)}} & 56.3\%\\
EOP & \makecell[r]{0.044\\\czp{(0.392-0.348)}} & \makecell[r]{0.160\\\czp{(0.721-0.562)}} & 62.5\% & \makecell[r]{0.025\\\czp{(0.631-0.605)}} & \makecell[r]{0.105\\\czp{(0.826-0.721)}} & 40.6\% & \makecell[r]{0.057\\\czp{(0.287-0.229)}} & \makecell[r]{0.221\\\czp{(0.612-0.391)}} & 68.8\%\\
FairSMOTE & \makecell[r]{0.049\\\czp{(0.397-0.348)}} & \makecell[r]{0.250\\\czp{(0.856-0.606)}} &75.0\% & \makecell[r]{0.081\\\czp{(0.687-0.605)}} & \makecell[r]{0.310\\\czp{(0.685-0.375)}} & 62.5\% &\makecell[r]{0.050\\\czp{(0.279-0.229)}} & \makecell[r]{0.312\\\czp{(0.745-0.433)}} & 68.8\%\\
LTDD & \makecell[r]{0.031\\\czp{(0.379-0.348)}} & \makecell[r]{0.182\\\czp{(0.788-0.606)}} &  53.1\% & \makecell[r]{0.044\\\czp{(0.649-0.605)}} & \makecell[r]{0.205\\\czp{(0.725-0.520)}} &  43.8\% & \makecell[r]{0.034\\\czp{(0.263-0.229)}} & \makecell[r]{0.213\\\czp{(0.646-0.433)}} & 46.9\%\\
MAAT & \makecell[r]{0.031\\\czp{(0.379-0.348)}} & \makecell[r]{0.151\\\czp{(0.755-0.605)}} & 62.5\% & \makecell[r]{0.035\\\czp{(0.640-0.605)}} & \makecell[r]{0.130\\\czp{(0.865-0.736)}} & 46.9\% & \makecell[r]{0.035\\\czp{(0.264-0.229)}} & \makecell[r]{0.178\\\czp{(0.610-0.433)}} & 59.4\%\\
FairMask & \makecell[r]{0.014\\\czp{(0.362-0.348)}} & \makecell[r]{0.060\\\czp{(0.622-0.562)}} & 37.5\% & \makecell[r]{0.015\\\czp{(0.621-0.605)}} & \makecell[r]{0.064\\\czp{(0.785-0.721)}} & 21.9\% & \makecell[r]{0.016\\\czp{(0.246-0.229)}} & \makecell[r]{0.101\\\czp{(0.611-0.509)}} & 25.0\%\\
MirrorFair & \makecell[r]{0.063\\\czp{(0.411-0.348)}} & \makecell[r]{0.241\\\czp{(0.846-0.605)}} & 84.4\% & \makecell[r]{0.072\\\czp{(0.677-0.605)}} & \makecell[r]{0.208\\\czp{(0.944-0.736)}} & 87.5\% & \makecell[r]{0.071\\\czp{(0.300-0.229)}} & \makecell[r]{0.286\\\czp{(0.718-0.433)}} & 81.3\%\\
\hline
\end{tabular}
\end{table*}

\noindent \textbf{Results.} Table \ref{effectsizetable} presents the results. The table presents the mean impact (the ``Mean'' columns) and the maximum decrease/increase in performance for the privileged/unprivileged group (the ``Max $\downarrow$'' and ``Max $\uparrow$'' columns), \czp{along with how these relative changes are derived from absolute numbers. For example, the mean impact of ADV in $SR_P$ is -0.026 (0.462-0.489), indicating that ADV reduces the average $SR_P$ across tasks from 0.489 to 0.462.} The table also shows the proportions of tasks where each method results in a significant decrease/increase in performance by a large effect size for the privileged/unprivileged groups (the ``Large $\downarrow$'' and ``Large $\uparrow$'' columns).

First, we find that existing methods substantially increase the benefits for the unprivileged group. Specifically, these methods significantly increase $SR_U$, $TPR_U$, and $FPR_U$ by a large effect size in 37.5\%$\sim$84.4\%, 21.9\%$\sim$87.5\%, and 25.0\%$\sim$81.3\% of tasks, respectively. In terms of the proportion of tasks with a significant increase in benefits for the unprivileged group, MirrorFair ranks first, followed by FairSMOTE.

Second, we observe that existing bias mitigation methods generally reduce the favorable outcomes for the privileged group, with significant decreases in $SR_P$, $TPR_P$, and $FPR_P$ by a large effect size in 31.3\%$\sim$62.5\%, 25.0\%$\sim$62.5\%, and 21.9\%$\sim$56.3\% of tasks, respectively. 

While these methods are effective in narrowing disparities, the reduction in favorable outcomes for the privileged group may raise concerns regarding social perceptions. Research suggests that members of the privileged group may perceive fairness efforts as introducing bias against them \cite{brown2022if, norton2011whites}, and thus resist fairness policies. If bias mitigation methods are perceived as disproportionately affecting the privileged group, this could result in resistance to their broader adoption, potentially complicating efforts to achieve widespread fairness. Addressing this challenge will be essential in promoting the successful implementation of bias mitigation strategies.

\finding{Existing bias mitigation methods substantially increase the benefits for the unprivileged group, with significant increases in the selection rate and true positive rate by a large effect size in 37.5\%$\sim$84.4\% and 21.9\%$\sim$87.5\% of tasks, respectively. However, these methods also lead to significant reductions in the selection rate and true positive rate for the privileged group, with large effect sizes observed in 31.3\%$\sim$62.5\% and 25.0\%$\sim$62.5\% of tasks, respectively. While these reductions aim to narrow disparities and promote fairness, they may also contribute to the perception that efforts to reduce bias against the unprivileged group introduce bias against the privileged group. Such perceptions could create tension and potentially hinder the wider adoption of bias mitigation strategies, complicating efforts to achieve equitable outcomes.}

\section{RQ2: Correlation Between Group Impact and Fairness}
\noindent \textbf{Methodology.} To explore the relationship between the impact on each group and fairness, we first calculate the changes in values for $SR_P$, $TPR_P$, $FPR_P$, $SR_U$, $TPR_U$, $FPR_U$, SPD, EOD, and AOD for each (bias mitigation method, task) pair by subtracting the original value from the value after applying the method. This generates a list of 256 value changes for each of the nine metrics, corresponding to the eight methods applied across 32  single-attribute bias mitigation tasks.

We then compute Spearman's correlation coefficient~\cite{myers2013research} between these lists for each pair of metrics. The coefficient $\rho$ ranges from -1 to 1, where 1 indicates a perfect positive correlation, 0 indicates no correlation, and -1 indicates a perfect negative correlation. A correlation is considered statistically significant when the $p$-value is below 0.05 \cite{ZPICSE24}.

\begin{figure}[!t]
\begin{center}
\includegraphics[width=0.6\columnwidth]{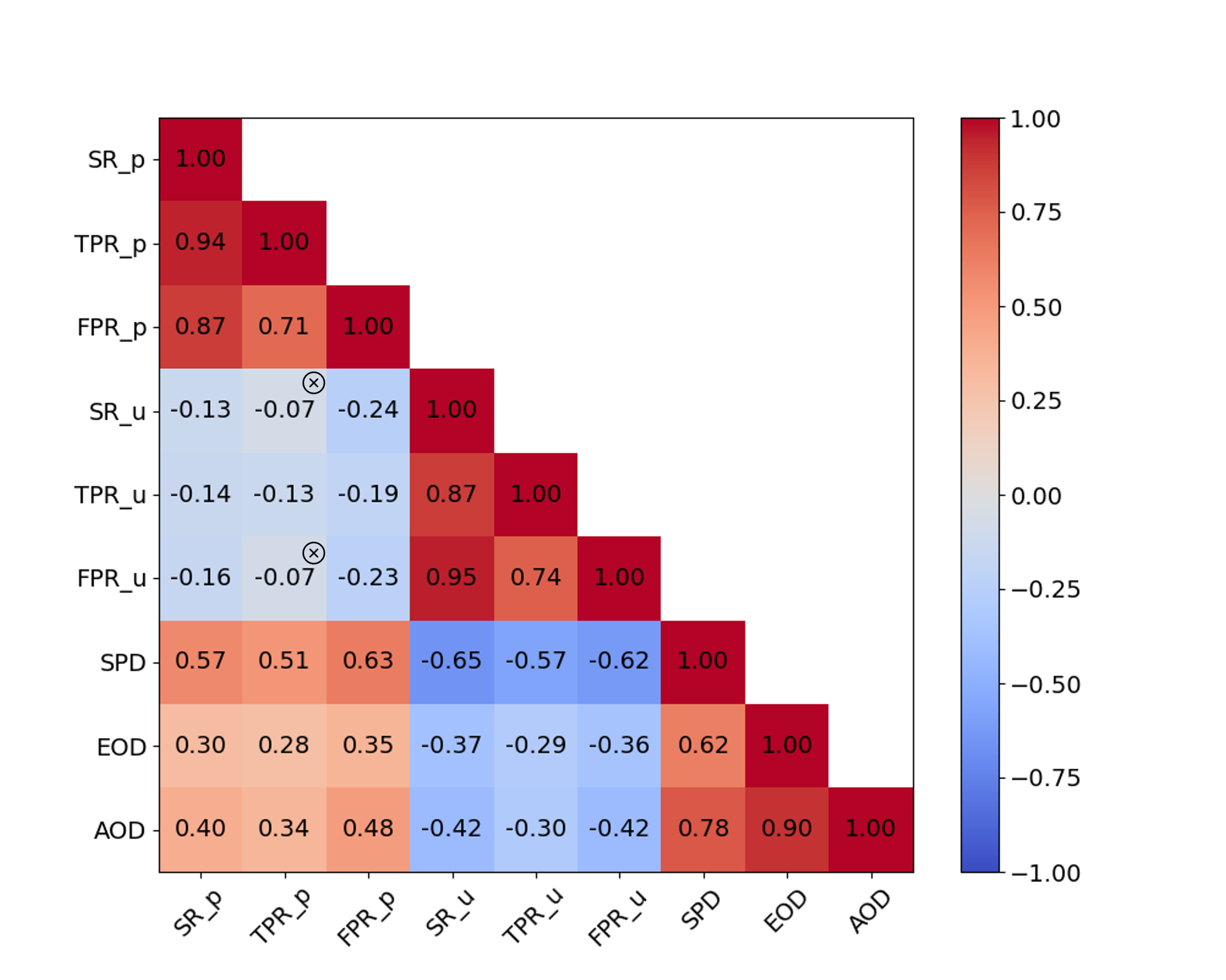}
\caption{(RQ2) Spearman's correlation across different metrics. The results reveal that greater fairness improvements, indicated by decreases in SPD, AOD, and EOD values, correspond to larger decreases in $SR_P$, $TPR_P$, and $FPR_P$, and larger increases in $SR_U$, $TPR_U$, and $FPR_U$.}\label{fig-correlation}
\end{center}
\end{figure}

\noindent \textbf{Results.} Figure \ref{fig-correlation} illustrates the correlation coefficients. All presented correlations are statistically significant except for the two cases marked with $\otimes$.

We observe that the three fairness metrics (i.e., SPD, EOD, and AOD) are positively correlated, consistent with previous empirical findings \cite{zeyujjj,Dabs220703277,biswas2020machine}, thereby validating the reliability of our correlation analysis. As a result, each fairness metric shows the same correlation trend with $SR_P$, $TPR_P$, $FPR_P$, $SR_U$, $TPR_U$, and $FPR_U$. Specifically, changes in SPD, EOD, and AOD are significantly positively correlated with $SR_P$, $TPR_P$, and $FPR_P$, and significantly negatively correlated with $SR_U$, $TPR_U$, and $FPR_U$. This indicates that greater fairness improvements, reflected by decreases in SPD, AOD, and EOD values, are associated with adjustments in SR, TPR, and FPR for both groups. In particular, fairness improvements are related to reductions in these metrics for the privileged group and corresponding increases for the unprivileged group, reflecting the narrowing of disparities.

\finding{In our study on bias mitigation methods for tabular data, greater fairness improvements are significantly associated with larger decreases in the selection rate, true positive rate, and false positive rate for the privileged group, and larger increases in these rates for the unprivileged group.}

\section{RQ3: Impact With Multiple Sensitive Attributes}
RQ3 explores whether the zero-sum pattern observed in RQ1 also applies to multiple-attribute tasks and examines how existing bias mitigation methods affect different demographic groups defined by multiple sensitive attributes.

\noindent \textbf{Methodology.} We consider 12 multiple-attribute tasks described in Section \ref{taskslist}. For each task, the population is divided into four demographic groups, as each of the two sensitive attributes creates a division between privileged and unprivileged groups. The four groups are ranked based on the proportion of members achieving favorable outcomes in the training data, resulting in $Group_1$, $Group_2$, $Group_3$, and $Group_4$. $Group_1$ represents the most  privileged group, while $Group_4$ represents the least  privileged. We employ the Mann-Whitney U-test, as described in Section \ref{rq1}, to analyze the impact of existing methods on each of the four groups. Since LTDD does not support handling multiple sensitive attributes, we consider the remaining seven bias mitigation methods for answering RQ3.

\noindent \textbf{Results.} Table \ref{frequencytable2} presents the results. Due to space constraints, we display only the number of tasks where each method significantly increases or decreases performance on each group. Overall, we find that the zero-sum pattern  also applies to multiple-attribute tasks.

First, bias mitigation methods tend to decrease the SR, TPR, and FPR for the most privileged group ($Group_1$). Specifically, these methods significantly reduce SR, TPR, and FPR for $Group_1$ in 48.8\% (41/84), 46.4\% (39/84), and 40.5\% (34/84) of tasks, respectively. Notably, only the EOP method significantly decreases FPR for $Group_1$ in 2 tasks, but it significantly increases FPR in 3 tasks. In all other scenarios, each method significantly decreases SR, TPR, and FPR for $Group_1$ in at least as many tasks as it increases them.

Second, these bias mitigation methods generally increase the SR, TPR, and FPR for the two least privileged groups ($Group_4$ and $Group_3$). Particularly, SR, TPR, and FPR for $Group_4$ are significantly increased in 67.9\% (57/84), 51.2\% (43/84), and 65.5\% (55/84) of tasks, respectively. Across all eight methods studied, SR, TPR, and FPR for $Group_4$ are significantly increased in more scenarios than they are decreased.

This finding contrasts with previous research~\cite{emnlp4BDK23}, which finds that bias mitigation methods in CV and NLP typically improve fairness by disadvantaging the most privileged group without benefiting the least privileged group when multiple sensitive attributes are considered.

Third, for the second privileged group ($Group_2$), the effects of bias mitigation methods are more balanced, with comparable increases and decreases in performance. For instance, SR is significantly increased in 23 tasks and decreased in 25 tasks. However, individual methods show varying trends; for example, FairMOTE is more likely to increase SR for $Group_2$, while MAAT tends to decrease it.

\finding{The zero-sum pattern of bias mitigation methods for tabular data also applies to multiple-attribute tasks. Unlike previous findings in CV and NLP, where methods often improve fairness by disadvantaging the most privileged group without benefiting the least privileged group in multiple-attributes tasks, our analysis reveals that methods for tabular data tend to decrease the selection rate, true positive rate, and false positive rate for the most privileged group, while increasing these metrics for the least privileged group.}

\begin{table*}[!tp]
\scriptsize
\centering
\caption{(RQ3) Number of tasks where each bias mitigation method significantly increases ($\uparrow$) or decreases ($\downarrow$) the SR, TPR, and FPR for each group when handling multiple sensitive attributes. Overall, bias mitigation methods for tabular data tend to decrease SR, TPR, and FPR for the most privileged group while increasing them for the least privileged group.}
\label{frequencytable2}
\begin{tabular}{l|rr|rr|rr|rr}
\hline
Method & $SR_1$ ($\uparrow$) & $SR_1$ ($\downarrow$) & $SR_2$ ($\uparrow$) & $SR_2$ ($\downarrow$) & $SR_3$ ($\uparrow$) & $SR_3$ ($\downarrow$)& $SR_4$ ($\uparrow$) & $SR_4$ ($\downarrow$)\\
\hline
ADV & 3 & 5 & 2 & 7 & 6 & 1 & 4 & 3\\
REW & 0 & 6 & 4 & 2 & 3 & 3 & 6 & 0\\
EOP & 2 & 3 & 4 & 1 & 7 & 0 & 8 & 0\\
FairSMOTE & 4 & 8 & 4 & 2 & 5 & 2 & 10 & 0\\
MAAT & 4 & 4 & 3 & 5 & 6 & 4 & 9 & 1\\
FairMask & 0 & 8 & 2 & 3 & 3 & 0 & 8 & 0\\
MirrorFair & 2 & 7 & 4 & 5 & 8 & 2 & 12 & 0\\
\hline
\rowcolor{gray!15}
Overall & 15 & 41 & 23 & 25 & 38 & 12 & 57 & 4\\
\hline
\hline
Method & $TPR_1$ ($\uparrow$) & $TPR_1$ ($\downarrow$) & $TPR_2$ ($\uparrow$) & $TPR_2$ ($\downarrow$) & $TPR_3$ ($\uparrow$) & $TPR_3$ ($\downarrow$)& $TPR_4$ ($\uparrow$) & $TPR_4$ ($\downarrow$)\\
\hline
ADV & 4 & 6 & 2 & 6 & 6 & 1 & 4 & 3\\
REW & 0 & 6 & 3 & 2 & 3 & 3 & 5 & 0\\
EOP & 2 & 3 & 3 & 2 & 5 & 1 & 3 & 0\\
FairSMOTE & 4 & 7 & 3 & 2 & 6 & 2 & 8 & 0\\
MAAT & 4 & 4 & 3 & 5 & 4 & 3 & 6 & 1\\
FairMask & 0 & 7 & 1 & 2 & 3 & 0 & 5 & 0\\
MirrorFair & 2 & 6 & 4 & 4 & 8 & 2 & 12 & 0\\
\hline
\rowcolor{gray!15}
Overall & 16 & 39 & 19 & 23 & 35 & 12 & 43 & 4\\
\hline
\hline
Method & $FPR_1$ ($\uparrow$) & $FPR_1$ ($\downarrow$) & $FPR_2$ ($\uparrow$) & $FPR_2$ ($\downarrow$) & $FPR_3$ ($\uparrow$) & $FPR_3$ ($\downarrow$)& $FPR_4$ ($\uparrow$) & $FPR_4$ ($\downarrow$)\\
\hline
ADV & 3 & 4 & 2 & 7 & 5 & 0 & 4 & 4\\
REW & 0 & 4 & 3 & 2 & 3 & 3 & 6 & 0\\
EOP & 3 & 2 & 5 & 0 & 6 & 0 & 8 & 0\\
FairSMOTE & 4 & 5 & 3 & 2 & 3 & 2 & 10 & 0\\
MAAT & 4 & 4 & 3 & 5 & 5 & 4 & 9 & 1\\
FairMask & 0 & 7 & 1 & 2 & 3 & 1 & 6 & 0\\
MirrorFair & 2 & 8 & 4 & 4 & 6 & 1 & 12 & 0\\
\hline
\rowcolor{gray!15}
Overall & 16 & 34 & 21 & 22 & 31 & 11 & 55 & 5 \\
\hline
\end{tabular}
\end{table*}

\section{RQ4: Bias Mitigation Without Harming Any Group}
Our investigation of the previous RQs shows that existing bias mitigation methods for tabular data often benefit the unprivileged group but at a significant cost to the privileged group. In this RQ, we explore whether it is possible to enhance the benefits for the unprivileged group without negatively impacting the privileged group. A straightforward approach is to apply bias mitigation methods exclusively to the unprivileged group while retaining the original predictions for the privileged group, given that current methods primarily benefit unprivileged groups.
We analyze the feasibility of this approach across both single-attribute tasks (\textbf{RQ4.1}) and multiple-attribute tasks (\textbf{RQ4.2}).

\subsection{RQ4.1: Without Harming Any Group in Single-Attribute Tasks}

\noindent \textbf{Methodology.} To implement this approach, we apply MirrorFair solely to the unprivileged group (denoted as \emph{MirrorFairU}), as MirrorFair has been shown to outperform other methods and is considered state-of-the-art \cite{xiao2024mirrorfair}. 
For comparison, we also implement a more aggressive approach aimed at equalizing the probability of receiving favorable outcomes (i.e., an equal selection rate) between the unprivileged and privileged groups. The implementation proceeds as follows: We randomly select 20\% of the training data as a validation set and use the remaining 80\% to train the model. The model is then used to predict labels for the privileged group members and the probability of receiving a favorable outcome for the unprivileged group members in the validation set. Based on these predictions, we calculate the selection rate for the privileged group, denoted as $x\%$. Next, we identify the top $x\%$ of predicted probabilities in the unprivileged group, with the lowest probability in this range set as the threshold for the unprivileged group. In the test data, if an unprivileged instance’s predicted probability exceeds this threshold, it is assigned a favorable outcome; otherwise, it receives an unfavorable outcome. For privileged group instances, we retain the original model predictions. This method ensures that both groups achieve the same selection rate in the test data, and we refer to it as \emph{Naivebase}.

Since MirrorFair is the state-of-the-art bias mitigation method for tabular data, we compare it against MirrorFairU and NaiveBase. In addition to evaluating fairness and group-specific performance, we also assess overall ML performance for a comprehensive evaluation. ML performance is a critical functional requirement of ML software. To measure it, we follow previous studies~\cite{ZPICSE24,xiao2024mirrorfair,sigsoftChenZSH22} and use a set of five common metrics: accuracy, precision, recall, F1-score, and Matthews Correlation Coefficient (MCC). For precision, recall, and F1-score, we report macro-average values, as done in prior research~\cite{ZPICSE24,xiao2024mirrorfair,sigsoftChenZSH22}, to provide an overall comparison across favorable and unfavorable classes by averaging the results for both classes. MCC is chosen for its ability to handle imbalanced class distributions, which are prevalent in fairness research benchmark datasets \cite{ZPICSE24,sigsoftChenZSH22}. This addresses concerns that accuracy, though most widely used in fairness studies \cite{Dabs220703277}, may not adequately reflect performance in the presence of imbalanced class distributions \cite{sigsoftChenZSH22,ZPICSE24}.

To conduct this comparison, we use a win-tie-loss analysis and apply the Mann-Whitney U-test described in Section \ref{rq1} to ensure the statistical significance of the results. For accuracy, precision, recall, F1-score, MCC, SR, and TPR, higher values indicate better performance; for FPR, SPD, EOD, and AOD, lower values reflect better performance and fairness. We compare the three methods across 32 single-attribute tasks. In each task, if MirrorFairU or NaiveBase produces significantly better results than MirrorFair, it is labeled as a ``Win.'' If the results are significantly worse, it is marked as a ``Loss.'' If there is no statistically significant difference, the outcome is classified as a ``Tie.'' We then count the number of tasks where MirrorFairU or NaiveBase achieves a Win, Tie, or Loss.

Additionally, in certain tasks, such as recidivism prediction, overall selection rates may not be limited. However, in resource-constrained applications such as loan approvals, selection rates might be capped. Thus, we also compare the overall selection rates of these approaches to evaluate if they significantly increase resource demands.

\begin{table*}[!tp]
\scriptsize
\centering
\caption{(RQ4.1) Comparative analysis of MirrorFairU vs. MirrorFair and NaiveBase vs. MirrorFair across 32 single-attribute tasks. The win-tie-loss analysis shows that MirrorFairU improves $SR_P$, $TPR_P$, and $FPR_P$ compared to MirrorFair, while preserving $SR_U$, $TPR_U$, $FPR_U$ and achieving similar or better overall ML performance, though with reduced fairness. In contrast, NaiveBase offers greater benefits to the unprivileged group, but at the cost of significantly lower overall ML performance.}
\label{frequencytable3}
\begin{tabular}{l|rrr|rrr}
\hline
\multirow{2}*{Metric}&\multicolumn{3}{c|}{MirrorFairU}&\multicolumn{3}{c}{NaiveBase}\\
& Win & Tie & Loss & Win & Tie & Loss \\
\hline
$SR_P$ & 19 & 12 & 1 & 19 & 12 & 1\\
$TPR_P$ & 16 & 15 & 1 & 17 & 14 & 1\\
$FPR_P$ & 0 & 14 & 18 & 0 & 13 & 19\\
$SR_U$ & 1 & 31 & 0 & 24 & 6 & 2\\
$TPR_U$ & 1 & 31 & 0 & 22 & 9 & 1\\
$FPR_U$ & 0 & 31 & 1 & 1 & 7 & 24\\
\hline
Accuracy & 2 & 28 & 2 & 0 & 8 & 24\\
Recall & 7 & 22 & 3 & 12 & 8 & 12 \\
Precision & 0 & 26 & 6 & 1 & 13 & 18\\
F1-score & 7 & 22 & 3 & 7 & 9 & 16\\
MCC & 7 & 22 & 3 & 4 & 14 & 14\\
\hline
SPD & 1 & 15 & 16 & 19 & 9 & 4\\
EOD & 2 & 23 & 7 & 1 & 10 & 21 \\
AOD & 1 & 19 & 12 & 2 & 12 & 18\\
\hline
\end{tabular}
\end{table*}

\noindent \textbf{Results.} Table \ref{frequencytable3} presents the results of the win-tie-loss analysis. We then compare MirrorFair with  MirrorFairU and NaiveBase, respectively.

\noindent \emph{MirrorFairU vs. MirrorFair:} 
MirrorFairU breaks the zero-sum pattern by enhancing benefits for the unprivileged group without reducing those for the privileged group. Specifically, MirrorFairU achieves higher $SP_P$ and $TPR_P$ compared to MirrorFair, as it applies bias mitigation only to the unprivileged group. Both methods achieve similar performance for the unprivileged group. Despite both methods treating the unprivileged group the same way, they do not yield identical results across all 32 tasks. This is due to the inherent nondeterminism in ML, where repeated training runs can produce different outcomes \cite{PhamQWLRTYN20}. However, this nondeterminism has an impact on only one task, and does not affect our main findings.

In terms of ML performance, the win-tie-loss analysis reveals that MirrorFairU and MirrorFair achieve similar accuracy levels, and MirrorFairU outperforms MirrorFair in recall, F1-score, and MCC. However, MirrorFairU has lower precision, which is expected since recall and precision are often conflicting objectives~\cite{buckland1994relationship}. This is why researchers often rely on the F1-score, which balances these two metrics. Overall, MirrorFairU surpasses MirrorFair in F1-score, with significantly higher results in 7 tasks and lower results in 3 tasks.

Regarding fairness metrics, MirrorFairU underperforms compared to MirrorFair on SPD, EOD, and AOD. This is because MirrorFairU increases benefits for the unprivileged group to match those of MirrorFair without reducing the benefits for the privileged group. However, from a welfare economics perspective, MirrorFairU achieves a Pareto improvement, meaning it makes at least one group better off without making others worse off \cite{hansson2004welfare}. This result exposes a limitation in existing fairness metrics: they primarily focus on the relative performance gap between groups and fail to account for cases where one group’s benefits improve without negatively affecting the other. As a result, more sophisticated evaluation methods are needed to better balance fairness and group performance in these complex scenarios.

\czp{In addition, the notable number of ties in Table~\ref{frequencytable3} is expected, as our goal in RQ4 is to show that applying MirrorFair exclusively to unprivileged groups (i.e., MirrorFairU) increases benefits for privileged groups, while preserving those for unprivileged groups and maintaining comparable overall ML performance, relative to applying it to the entire population. As a result, ties are prevalent in metrics such as $SR_U$, $TPR_U$, and $FPR_U$, and overall ML performance metrics.}

\noindent \emph{NaiveBase vs. MirrorFair:} As shown in Table \ref{frequencytable3}, NaiveBase provides greater benefits for both privileged and unprivileged groups, with higher $SP_P$ in 19 tasks and higher $SP_U$ in 24 tasks. However, this advantage comes with a significant cost: NaiveBase exhibits considerably lower accuracy than MirrorFair in 24 out of 32 tasks. Given the critical importance of ML performance for practical applications, NaiveBase's reduced accuracy makes it less feasible for real-world use. Therefore, we do not consider it further in our analysis.

Based on our analysis, MirrorFairU shows promise as a bias mitigation method. To further assess its practicality, we calculate its overall selection rate to ensure it does not impose excessive resource demands. We compare MirrorFairU to MirrorFair, the current state-of-the-art method for bias mitigation in tabular data. Our findings indicate that MirrorFairU leads to an average increase of 0.033 in the overall selection rate across all 32 tasks, compared to 0.023 for MirrorFair. This translates to a marginal increase of about 1\% in resource requirements when using MirrorFairU instead of MirrorFair. Moreover, in tasks such as recidivism prediction, this increase in the selection rate does not demand additional social resources.

\finding{Applying the state-of-the-art bias mitigation method, MirrorFair, exclusively to the unprivileged group breaks the zero-sum pattern by increasing benefits for the unprivileged group without diminishing those for the privileged group. This approach not only enhances fairness and improves outcomes for the unprivileged group, but also preserves the benefits for the privileged group while maintaining similar or even better overall ML performance compared to applying MirrorFair to the entire population. Additionally, this approach leads to an average increase of only 0.01 in the overall selection rate.}

\subsection{RQ4.2: Without Harming Any Group in Multiple-Attribute Tasks}
\noindent \textbf{Methodology.} To further assess the applicability of MirrorFairU, we apply it to 12 multiple-attribute tasks. As shown in Table~\ref{frequencytable2}, MirrorFair primarily negatively impacts $Group_1$ and $Group_2$ while benefiting $Group_3$ and $Group_4$. Therefore, for the multiple-attribute tasks, we implement MirrorFairU by applying MirrorFair exclusively to $Group_3$ and $Group_4$. We then evaluate MirrorFairU’s fairness, performance, and overall selection rates using the same methodology as in RQ4.1.

\noindent \textbf{Results.} Table \ref{frequencytable4} presents the results, which generally align with our findings in RQ4.1. MirrorFairU achieves comparable performance for $Group_3$ and $Group_4$ as MirrorFair, while improving the SR, TPR, and FPR for $Group_1$ and $Group_2$, as expected, since MirrorFair is applied only to $Group_1$ and $Group_2$. In terms of overall performance, MirrorFairU exhibits similar accuracy, F1-score, and MCC compared to MirrorFair. Moreover, as noted in RQ4.1, MirrorFairU shows better recall but worse precision than MirrorFair. Additionally, consistent with our findings in RQ4.1, MirrorFairU demonstrates poorer fairness than MirrorFair regarding SPD, EOD, and AOD.

We also calculate the overall selection rate for both MirrorFairU and MirrorFair. MirrorFairU shows an average increase of 0.081 across all 12 multiple-attribute tasks, compared to 0.064 for MirrorFair, resulting in a difference of just 0.017.

\begin{table*}[!tp]
\scriptsize
\centering
\caption{(RQ4.2) Comparative analysis of MirrorFairU vs. MirrorFair across 12 multiple-attribute tasks. The win-tie-loss results indicate that MirrorFairU enhances benefits for $Group_1$ and $Group_2$ compared to MirrorFair, while maintaining the benefits for $Group_3$ and $Group_4$ and achieving similar overall ML performance.}
\label{frequencytable4}
\begin{tabular}{l|rrr||l|rrr}
\hline
\multirow{2}*{Metric}&\multicolumn{3}{c||}{MirrorFairU} & \multirow{2}*{Metric}&\multicolumn{3}{c}{MirrorFairU}\\
& Win & Tie & Loss &  & Win & Tie & Loss\\
\hline
$SR_1$ & 8 & 4 & 0& Accuracy & 2 & 9 & 1 \\
$TPR_1$ & 8 & 4 & 0& Recall & 3 & 8 & 1\\
$FPR_1$ & 0 & 3 & 9& Precision & 0 & 8 & 4\\
$SR_2$ & 6 & 5 & 1& F1-score & 2 & 8 & 2\\
$TPR_2$ & 4 & 7 & 1& MCC & 2 & 9 & 1\\
$FPR_2$ & 1 & 6 & 5& SPD & 1 & 5 &6 \\
$SR_3$ & 0 & 12 & 0& EOD & 0 & 10 & 2\\
$TPR_3$ & 0 & 12 & 0& AOD & 1 & 6 & 5\\
\cline{5-8}
$FPR_3$ & 0 & 12 & 0\\
$SR_4$ & 0& 12 & 0\\
$TPR_4$ & 0 & 12 & 0\\
$FPR_4$ & 0 & 12 & 0\\
\cline{1-4}
\end{tabular}
\end{table*}

\finding{Applying the state-of-the-art bias mitigation method, MirrorFair, exclusively to unprivileged groups in multi-attribute tasks increases benefits for privileged groups compared to the standard MirrorFair, while preserving benefits for unprivileged groups and maintaining similar ML performance. Additionally, the average difference in the overall selection rate between the two approaches is only 0.017.}

\section{Discussion}
This section explores the potential reason for the difference between our findings and prior research. It also discusses \czp{the qualitative insights into the observed zero-sum trade-offs}, the implications for various stakeholders, and potential threats to validity.

\subsection{Comparison With Previous Work}
Previous studies in CV and NLP have found that bias mitigation methods in these fields often lead to a leveling-down effect, where the performance of all groups is degraded \cite{emnlp4BDK23,ZietlowLBKLS022}. In contrast, our findings reveal that bias mitigation methods applied to tabular data exhibit a zero-sum pattern. In this case, the methods improve outcomes for the unprivileged group while reducing them for the privileged group. This pattern holds consistently across all eight bias mitigation methods we examine. In this section, we explore the possible reason for this difference.

Zietlow et al.~\cite{ZietlowLBKLS022} suggest that the leveling-down phenomenon in CV occurs due to the high-dimensional data and high-capacity models often used. These models typically achieve near-zero training error when fairness constraints are applied, but struggle to generalize to new test data, leading to degraded performance for all groups.

In our experiments, however, we observe an average training error of 9\% when applying the eight bias mitigation methods across 32 single-attribute tasks. Additionally, the performance impact for each group on the training data is consistent with the test data. Specifically, these methods increase the SR, TPR, and FPR for the unprivileged group while decreasing them for the privileged group. Due to space constraints, Table \ref{frequencytable5} presents only the SR metric, but similar patterns hold across other metrics. We find that all eight methods tend to decrease $SR_P$ and increase $SR_U$ on the training data. This suggests that bias mitigation methods for tabular data may generalize better from training to test data compared to those used in CV and NLP, leading to the observed differences.

\begin{table*}[!tp]
\scriptsize
\centering
\caption{Number of tasks where each bias mitigation method significantly increases ($\uparrow$), decreases ($\downarrow$), or does not significantly impact ($-$) the SR for each group in the training data. Overall, when applying bias mitigation methods for tabular data to the training data, these methods tend to decrease SR for the privileged group while increasing it for the unprivileged group.}
\label{frequencytable5}
\begin{tabular}{l|rrr|rrr}
\hline
Method & $SR_P$ ($\uparrow$) & $SR_P$ ($-$) & $SR_P$ ($\downarrow$) & $SR_U$ ($\uparrow$) & $SR_U$ ($-$) & $SR_U$ ($\downarrow$) \\
\hline
ADV & 6 & 9 & 17 & 13 & 13 & 6\\
REW & 0 & 7 & 25 & 24 & 8 & 0\\
EOP & 1 & 12 & 19 & 22 & 10 & 0\\
FairSMOTE & 8 & 9 & 15 & 24 & 4 & 4\\
LTDD & 1 & 15 & 16 & 18 & 14 & 0\\
MAAT & 2 & 9 & 21 & 27 & 5 & 0\\
FairMask & 0 & 18 & 14 & 12 & 20 & 0\\
MirrorFair & 4 & 11 & 17 & 29 & 1 & 2 \\
\hline
\rowcolor{gray!15}
Overall & 22 & 90 & 144 & 169 & 75 & 12\\
\hline
\end{tabular}
\end{table*}

\czp{\subsection{Qualitative Insights}
We provide qualitative insights into the mechanisms of each bias mitigation method to explain why they lead to observed trade-offs between privileged and unprivileged groups. \textbf{(1) ADV} reduces the model’s reliance on sensitive attributes and their proxies, suppressing correlations that disproportionately benefit privileged groups. This limits the predictive advantage of features tied to privilege, decreasing their benefits. Unprivileged groups gain as the model learns to rely on features more equitably distributed across demographics, improving their outcomes. \textbf{(2) REW} adjusts instance weights to prioritize unprivileged groups, focusing the model’s learning on improving their outcomes. This rebalancing reduces the influence of privileged group instances, decreasing their benefits while enhancing those of unprivileged groups. \textbf{(3) EOP} adjusts predictions to equalize false positive and false negative rates across groups. These adjustments often reduce accuracy for privileged groups by correcting biases that previously favor them. Unprivileged groups benefit as their predictive outcomes become more equitable. \textbf{(4) FairSMOTE} generates synthetic data for unprivileged groups, increasing their representation in training. This improves the model's ability to generalize for unprivileged groups but reduces the relative influence of privileged group data, decreasing their benefits. \textbf{(5) LTDD} removes biased components of training data that favor privileged groups, reducing the model's reliance on features tied to privilege. This levels the playing field, improving outcomes for unprivileged groups but decreasing benefits for privileged ones. \textbf{(6)~MAAT} combines predictions from models optimized separately for fairness and performance. The fairness-optimized model redistributes benefits to reduce disparities, often at the cost of benefits for privileged groups. Unprivileged groups gain as their outcomes are specifically targeted for improvement. \textbf{(7) FairMask} predicts sensitive attributes from other features and relabels them to reduce bias. By masking biased correlations that favor privileged groups, the model redistributes benefits, improving outcomes for unprivileged groups while reducing those of privileged groups. \textbf{(8) MirrorFair} uses counterfactual datasets to balance predictions, reducing the model's reliance on patterns that favor privileged groups. This adjustment corrects disadvantages for unprivileged groups but decreases the benefits for privileged ones as outcomes are redistributed.
}

\subsection{Implications}
\czp{
\noindent \textbf{\emph{For SE researchers:}} \textbf{(1) Research reproducibility.} Fairness research findings in CV and NLP may not generalize to tabular data, as shown in our analysis (RQ1 and RQ3). This underscores the need for domain-specific evaluations of bias mitigation methods. SE researchers should replicate and validate software fairness studies across data types and task domains to identify limitations and ensure robust outcomes in diverse ML applications. \textbf{(2) Comprehensive evaluation practices.} Current software fairness research primarily focuses on overall performance and fairness, often overlooking the performance of individual demographic groups. Our findings (RQs1–3) show significant trade-offs in group benefits, which could hinder the practical adoption of bias mitigation methods. SE researchers should adopt granular evaluation practices that assess impacts on each demographic group, highlighting trade-offs and enabling more effective interventions that balance fairness and group-specific outcomes. \textbf{(3) Rethinking fairness metrics.} Our findings (RQ4) show that while MirrorFairU provides comparable benefits for unprivileged groups and better outcomes for privileged groups, existing fairness metrics still favor MirrorFair by focusing on relative disparities and often overlooking the absolute performance of each group. This limitation can obscure opportunities to improve outcomes for all groups. Fairness metrics should evolve to balance equity across groups with absolute benefits for each, considering both relative disparities and individual group outcomes to promote fairness and well-being for all demographics. \textbf{(4) Innovative bias mitigation methods.} The trade-offs between groups in existing methods present challenges in high-stakes applications like criminal justice, healthcare, and finance, where the performance of all groups must remain high. Our results (RQ4) demonstrate the feasibility of achieving fairness without such trade-offs, providing a viable path forward. SE researchers should develop bias mitigation methods that optimize both relative fairness and absolute performance across groups. Prioritizing these solutions can lead to interventions that are ethical, effective, and more likely to be adopted in real-world systems.

\noindent \textbf{\emph{For software engineers and practitioners:}} \textbf{(1) Mediating trade-offs and addressing conflicting requirements.} Anti-discrimination laws require software engineers and practitioners to ensure fairness in ML systems. However, our findings (RQs1–3) reveal that fairness interventions often involve trade-offs, where improvements for unprivileged groups significantly reduce outcomes for privileged groups, especially as fairness gains increase. These trade-offs can hinder the adoption of software systems. Engineers and practitioners should consider these conflicting requirements in the SE process, using negotiation, mediation, conflict resolution, and multi-objective optimization strategies to balance these requirements. \textbf{(2) Leveraging evidence-based practices.} Our evaluation of eight advanced bias mitigation methods (RQ1 and RQ3) highlights their impacts on each group’s performance across single and multiple sensitive attribute scenarios, providing guidance for selecting methods aligned with specific group performance goals. Additionally, our results (RQ4) propose a promising approach: selectively applying bias mitigation to enhance fairness and unprivileged group outcomes while preserving privileged group benefits. This strategy offers a practical path for achieving balanced outcomes, enabling engineers and practitioners to develop fair and effective ML systems.

\noindent \textbf{\emph{For policymakers:}} \textbf{(1) Establishing fairness standards linked to group-specific performance.} Current regulations aim to promote fairness in software systems, yet our findings (RQ1 and RQ3) show that bias mitigation methods often improve fairness at the expense of privileged groups, with greater fairness gains correlating with larger reductions in their outcomes (RQ2). These trade-offs could hinder the adoption of fairness policies. Policymakers should establish standards requiring detailed reporting of fairness metrics along with their impact on all demographic groups, encouraging balanced approaches that consider the broader implications of fairness interventions. \textbf{(2) Challenging the zero-sum perception of fairness.} Our findings (RQ4) demonstrate that fairness improvements for underprivileged groups do not always demand significant sacrifices from privileged groups. Policymakers should counter the misconception of fairness as a zero-sum game through awareness campaigns and stakeholder dialogues. Presenting evidence that fairness can be achieved without severe trade-offs will build trust and support for fairness-focused regulations and fair software systems.}

\subsection{Threats to Validity}
\noindent \textbf{\emph{Selection of bias mitigation tasks.}} The choice of tasks may pose a threat to validity. To address this, we employ the same set of 44 bias mitigation tasks from a recent study \cite{xiao2024mirrorfair}, covering five widely used datasets and four commonly applied ML models. These datasets reflect real-world applications across various domains and include sensitive attributes frequently considered in practice. \czp{Analyzing SE-specific datasets could further strengthen the relevance of this work. However, we are unaware of any publicly available SE tabular datasets suitable for such fairness studies. This may explain why recent related papers~\cite{biswas2020machine,fairsmotepaper,sigsoftChenZSH22,Dabs220703277,ZPICSE24,sigsoftHortZSH21,icseLiMC0WZX22,fairmaskpaper,xiao2024mirrorfair,icseZhangH21} also rely on datasets from other domains.}

\noindent \textbf{\emph{Selection of bias mitigation methods.}} Given the extensive volume of research on bias mitigation~\cite{DBcorrabs220707068}, it is challenging to include all existing methods in this study. To mitigate this threat, we select a set of eight representative bias mitigation methods, encompassing both the most widely used methods and recently proposed state-of-the-art techniques.

\noindent \textbf{\emph{Selection of statistical analysis methods.}} We employ the Mann-Whitney U-test, Cliff's $\delta$, and Spearman's correlation coefficient for our statistical analysis. These methods are commonly used in the SE literature \cite{ZPICSE24, Dabs220703277}. They do not rely on the assumption of data normality, making them appropriate for our study, which involves various data that may not conform to a normal distribution. 

\noindent \textbf{\emph{Selection of metrics.}} To mitigate the potential threat related to metric selection, we adopt the same set of metrics used in recent software fairness studies \cite{xiao2024mirrorfair, ZPICSE24, sigsoftChenZSH22}. This includes three widely recognized fairness metrics and five commonly used ML performance metrics.

\noindent \textbf{\emph{Implementation of bias mitigation methods.}} To address this threat, we directly use the code provided by the original authors of the bias mitigation methods, ensuring implementation reliability. Additionally, in alignment with prior work \cite{ZPICSE24}, each method is applied to each task 20 times to mitigate the influence of randomness. All code and datasets used in this study have been made available in an open repository \cite{githublink} to facilitate replication and validation.

\section{Conclusion}
This paper presents a comprehensive study on the impact of existing bias mitigation methods for tabular data, analyzing their effects on the model's performance for different demographic groups. We observe that all methods lead to a zero-sum trade-off, where improvements for unprivileged groups are associated with reduced outcomes for privileged groups. Given that the perception of such trade-offs might impede the broader adoption of fairness-focused policies, we explore the application of the state-of-the-art bias mitigation method exclusively to unprivileged groups. Our preliminary findings suggest that this approach can enhance benefits for unprivileged groups without compromising outcomes for privileged groups, and it maintains overall ML performance (e.g., accuracy and F1-score) comparable to existing methods. Building on these insights, we plan to develop more effective bias mitigation strategies that avoid the zero-sum trade-off, thereby facilitating the broader adoption of bias mitigation methods in ML software.

\section{Data Availability}
We have made the replication package publicly available in a repository \cite{githublink}, which includes all datasets, code, and intermediate results from our study.

\section*{Acknowledgments}
This research is supported by the National Research Foundation Singapore and DSO National Laboratories under the AI Singapore Programme (AISG Award No. AISG2-RP-2020-019); by the National Research Foundation Singapore and the Cyber Security Agency of Singapore under the National Cybersecurity R\&D Programme (NCRP25-P04-TAICeN); and by the National Research Foundation, Prime Minister’s Office, Singapore under the Campus for Research Excellence and Technological Enterprise (CREATE) programme. Any opinions, findings, conclusions, or recommendations expressed in this paper are those of the authors and do not reflect the views of the National Research Foundation Singapore or the Cyber Security Agency of Singapore.

\bibliographystyle{ACM-Reference-Format}
\bibliography{fairnessbib}


\begin{thebibliography}{43}


\ifx \showCODEN    \undefined \def \showCODEN     #1{\unskip}     \fi
\ifx \showISBNx    \undefined \def \showISBNx     #1{\unskip}     \fi
\ifx \showISBNxiii \undefined \def \showISBNxiii  #1{\unskip}     \fi
\ifx \showISSN     \undefined \def \showISSN      #1{\unskip}     \fi
\ifx \showLCCN     \undefined \def \showLCCN      #1{\unskip}     \fi
\ifx \shownote     \undefined \def \shownote      #1{#1}          \fi
\ifx \showarticletitle \undefined \def \showarticletitle #1{#1}   \fi
\ifx \showURL      \undefined \def \showURL       {\relax}        \fi
\providecommand\bibfield[2]{#2}
\providecommand\bibinfo[2]{#2}
\providecommand\natexlab[1]{#1}
\providecommand\showeprint[2][]{arXiv:#2}

\bibitem[git(2025)]%
        {githublink}
 \bibinfo{year}{2025}\natexlab{}.
\newblock \bibinfo{title}{Replication package}.
\newblock \bibinfo{howpublished}{\url{https://github.com/chenzhenpeng18/FSE25-ZeroSum}}.
\newblock


\bibitem[Alvarez and Menzies(2023)]%
        {softwareAlvarezM23}
\bibfield{author}{\bibinfo{person}{Lauren Alvarez} {and} \bibinfo{person}{Tim Menzies}.} \bibinfo{year}{2023}\natexlab{}.
\newblock \showarticletitle{Don't lie to me: Avoiding malicious explanations with {STEALTH}}.
\newblock \bibinfo{journal}{\emph{{IEEE} Softwaware}} \bibinfo{volume}{40}, \bibinfo{number}{3} (\bibinfo{year}{2023}), \bibinfo{pages}{43--53}.
\newblock


\bibitem[Bennin et~al\mbox{.}(2017)]%
        {esemBenninKMPM17}
\bibfield{author}{\bibinfo{person}{Kwabena~Ebo Bennin}, \bibinfo{person}{Jacky Keung}, \bibinfo{person}{Akito Monden}, \bibinfo{person}{Passakorn Phannachitta}, {and} \bibinfo{person}{Solomon Mensah}.} \bibinfo{year}{2017}\natexlab{}.
\newblock \showarticletitle{The significant effects of data sampling approaches on software defect prioritization and classification}. In \bibinfo{booktitle}{\emph{Proceedings of the 2017 {ACM/IEEE} International Symposium on Empirical Software Engineering and Measurement, {ESEM} 2017}}. \bibinfo{pages}{364--373}.
\newblock


\bibitem[Biswas and Rajan(2020)]%
        {biswas2020machine}
\bibfield{author}{\bibinfo{person}{Sumon Biswas} {and} \bibinfo{person}{Hridesh Rajan}.} \bibinfo{year}{2020}\natexlab{}.
\newblock \showarticletitle{Do the machine learning models on a crowd sourced platform exhibit bias? An empirical study on model fairness}. In \bibinfo{booktitle}{\emph{Proceedings of the 28th ACM Joint Meeting on European Software Engineering Conference and Symposium on the Foundations of Software Engineering, ESEC/FSE 2020}}. \bibinfo{pages}{642--653}.
\newblock


\bibitem[Biswas and Rajan(2021)]%
        {sigsoftBiswasR21}
\bibfield{author}{\bibinfo{person}{Sumon Biswas} {and} \bibinfo{person}{Hridesh Rajan}.} \bibinfo{year}{2021}\natexlab{}.
\newblock \showarticletitle{Fair preprocessing: Towards understanding compositional fairness of data transformers in machine learning pipeline}. In \bibinfo{booktitle}{\emph{Proceedings of the 29th {ACM} Joint European Software Engineering Conference and Symposium on the Foundations of Software Engineering, {ESEC/FSE} 2021}}. \bibinfo{pages}{981--993}.
\newblock


\bibitem[Brown et~al\mbox{.}(2022)]%
        {brown2022if}
\bibfield{author}{\bibinfo{person}{N~Derek Brown}, \bibinfo{person}{Drew~S Jacoby-Senghor}, {and} \bibinfo{person}{Isaac Raymundo}.} \bibinfo{year}{2022}\natexlab{}.
\newblock \showarticletitle{If you rise, I fall: Equality is prevented by the misperception that it harms advantaged groups}.
\newblock \bibinfo{journal}{\emph{Science Advances}} \bibinfo{volume}{8}, \bibinfo{number}{18} (\bibinfo{year}{2022}), \bibinfo{pages}{eabm2385}.
\newblock


\bibitem[Brun and Meliou(2018)]%
        {sigsoftBrunM18}
\bibfield{author}{\bibinfo{person}{Yuriy Brun} {and} \bibinfo{person}{Alexandra Meliou}.} \bibinfo{year}{2018}\natexlab{}.
\newblock \showarticletitle{Software fairness}. In \bibinfo{booktitle}{\emph{Proceedings of the 2018 {ACM} Joint Meeting on European Software Engineering Conference and Symposium on the Foundations of Software Engineering, {ESEC/FSE} 2018}}. \bibinfo{pages}{754--759}.
\newblock


\bibitem[Buckland and Gey(1994)]%
        {buckland1994relationship}
\bibfield{author}{\bibinfo{person}{Michael Buckland} {and} \bibinfo{person}{Fredric Gey}.} \bibinfo{year}{1994}\natexlab{}.
\newblock \showarticletitle{The relationship between recall and precision}.
\newblock \bibinfo{journal}{\emph{Journal of the American Society for Information Science}} \bibinfo{volume}{45}, \bibinfo{number}{1} (\bibinfo{year}{1994}), \bibinfo{pages}{12--19}.
\newblock


\bibitem[Chakraborty et~al\mbox{.}(2021)]%
        {fairsmotepaper}
\bibfield{author}{\bibinfo{person}{Joymallya Chakraborty}, \bibinfo{person}{Suvodeep Majumder}, {and} \bibinfo{person}{Tim Menzies}.} \bibinfo{year}{2021}\natexlab{}.
\newblock \showarticletitle{Bias in machine learning software: Why? How? What to do?}. In \bibinfo{booktitle}{\emph{Proceedings of the 29th {ACM} Joint European Software Engineering Conference and Symposium on the Foundations of Software Engineering, {ESEC/FSE} 2021}}. \bibinfo{pages}{429--440}.
\newblock


\bibitem[Chakraborty et~al\mbox{.}(2020)]%
        {fairwaypaper}
\bibfield{author}{\bibinfo{person}{Joymallya Chakraborty}, \bibinfo{person}{Suvodeep Majumder}, \bibinfo{person}{Zhe Yu}, {and} \bibinfo{person}{Tim Menzies}.} \bibinfo{year}{2020}\natexlab{}.
\newblock \showarticletitle{Fairway: A way to build fair {ML} software}. In \bibinfo{booktitle}{\emph{Proceedings of the 28th {ACM} Joint European Software Engineering Conference and Symposium on the Foundations of Software Engineering, {ESEC/FSE} 2020}}. \bibinfo{pages}{654--665}.
\newblock


\bibitem[Chen et~al\mbox{.}(2024a)]%
        {Dabs220710223}
\bibfield{author}{\bibinfo{person}{Zhenpeng Chen}, \bibinfo{person}{Jie~M. Zhang}, \bibinfo{person}{Max Hort}, \bibinfo{person}{Mark Harman}, {and} \bibinfo{person}{Federica Sarro}.} \bibinfo{year}{2024}\natexlab{a}.
\newblock \showarticletitle{Fairness testing: A comprehensive survey and analysis of trends}.
\newblock \bibinfo{journal}{\emph{ACM Transactions on Software Engineering and Methodology}} \bibinfo{volume}{33}, \bibinfo{number}{5} (\bibinfo{year}{2024}), \bibinfo{pages}{137:1--137:59}.
\newblock


\bibitem[Chen et~al\mbox{.}(2022)]%
        {sigsoftChenZSH22}
\bibfield{author}{\bibinfo{person}{Zhenpeng Chen}, \bibinfo{person}{Jie~M. Zhang}, \bibinfo{person}{Federica Sarro}, {and} \bibinfo{person}{Mark Harman}.} \bibinfo{year}{2022}\natexlab{}.
\newblock \showarticletitle{{MAAT:} A novel ensemble approach to addressing fairness and performance bugs for machine learning software}. In \bibinfo{booktitle}{\emph{Proceedings of the 30th {ACM} Joint European Software Engineering Conference and Symposium on the Foundations of Software Engineering, {ESEC/FSE} 2022}}. \bibinfo{pages}{1122--1134}.
\newblock


\bibitem[Chen et~al\mbox{.}(2023)]%
        {Dabs220703277}
\bibfield{author}{\bibinfo{person}{Zhenpeng Chen}, \bibinfo{person}{Jie~M. Zhang}, \bibinfo{person}{Federica Sarro}, {and} \bibinfo{person}{Mark Harman}.} \bibinfo{year}{2023}\natexlab{}.
\newblock \showarticletitle{A comprehensive empirical study of bias mitigation methods for machine learning classifiers}.
\newblock \bibinfo{journal}{\emph{{ACM} Transactions on Software Engineering and Methodology}} \bibinfo{volume}{32}, \bibinfo{number}{4} (\bibinfo{year}{2023}), \bibinfo{pages}{106:1--106:30}.
\newblock


\bibitem[Chen et~al\mbox{.}(2024b)]%
        {ZPICSE24}
\bibfield{author}{\bibinfo{person}{Zhenpeng Chen}, \bibinfo{person}{Jie~M. Zhang}, \bibinfo{person}{Federica Sarro}, {and} \bibinfo{person}{Mark Harman}.} \bibinfo{year}{2024}\natexlab{b}.
\newblock \showarticletitle{Fairness improvement with multiple protected attributes: How far are we?}. In \bibinfo{booktitle}{\emph{Proceedings of the 46th {IEEE/ACM} International Conference on Software Engineering, {ICSE} 2024}}. \bibinfo{pages}{160:1--160:13}.
\newblock


\bibitem[Friedler et~al\mbox{.}(2019)]%
        {FriedlerSVCHR19}
\bibfield{author}{\bibinfo{person}{Sorelle~A. Friedler}, \bibinfo{person}{Carlos Scheidegger}, \bibinfo{person}{Suresh Venkatasubramanian}, \bibinfo{person}{Sonam Choudhary}, \bibinfo{person}{Evan~P. Hamilton}, {and} \bibinfo{person}{Derek Roth}.} \bibinfo{year}{2019}\natexlab{}.
\newblock \showarticletitle{A comparative study of fairness-enhancing interventions in machine learning}. In \bibinfo{booktitle}{\emph{Proceedings of the Conference on Fairness, Accountability, and Transparency, FAT 2019}}. \bibinfo{pages}{329--338}.
\newblock


\bibitem[Gao et~al\mbox{.}(2022)]%
        {icseGaoZMSCW22}
\bibfield{author}{\bibinfo{person}{Xuanqi Gao}, \bibinfo{person}{Juan Zhai}, \bibinfo{person}{Shiqing Ma}, \bibinfo{person}{Chao Shen}, \bibinfo{person}{Yufei Chen}, {and} \bibinfo{person}{Qian Wang}.} \bibinfo{year}{2022}\natexlab{}.
\newblock \showarticletitle{Fairneuron: Improving deep neural network fairness with adversary games on selective neurons}. In \bibinfo{booktitle}{\emph{Proceedings of the 44th {IEEE/ACM} International Conference on Software Engineering, {ICSE} 2022}}. \bibinfo{pages}{921--933}.
\newblock


\bibitem[Hansson(2004)]%
        {hansson2004welfare}
\bibfield{author}{\bibinfo{person}{Sven~Ove Hansson}.} \bibinfo{year}{2004}\natexlab{}.
\newblock \showarticletitle{Welfare, justice, and Pareto efficiency}.
\newblock \bibinfo{journal}{\emph{Ethical Theory and Moral Practice}}  \bibinfo{volume}{7} (\bibinfo{year}{2004}), \bibinfo{pages}{361--380}.
\newblock


\bibitem[Hardt et~al\mbox{.}(2016)]%
        {EOpaper}
\bibfield{author}{\bibinfo{person}{Moritz Hardt}, \bibinfo{person}{Eric Price}, {and} \bibinfo{person}{Nati Srebro}.} \bibinfo{year}{2016}\natexlab{}.
\newblock \showarticletitle{Equality of opportunity in supervised learning}. In \bibinfo{booktitle}{\emph{Proceedings of the Annual Conference on Neural Information Processing Systems 2016, NIPS 2016}}. \bibinfo{pages}{3315--3323}.
\newblock


\bibitem[Horkoff(2019)]%
        {reHorkoff19}
\bibfield{author}{\bibinfo{person}{Jennifer Horkoff}.} \bibinfo{year}{2019}\natexlab{}.
\newblock \showarticletitle{Non-Functional requirements for machine learning: Challenges and new directions}. In \bibinfo{booktitle}{\emph{Proceedings of the 27th {IEEE} International Requirements Engineering Conference, {RE} 2019}}. \bibinfo{pages}{386--391}.
\newblock


\bibitem[Hort et~al\mbox{.}(2024)]%
        {DBcorrabs220707068}
\bibfield{author}{\bibinfo{person}{Max Hort}, \bibinfo{person}{Zhenpeng Chen}, \bibinfo{person}{Jie~M. Zhang}, \bibinfo{person}{Mark Harman}, {and} \bibinfo{person}{Federica Sarro}.} \bibinfo{year}{2024}\natexlab{}.
\newblock \showarticletitle{Bias mitigation for machine learning classifiers: {A} comprehensive survey}.
\newblock \bibinfo{journal}{\emph{ACM Journal on Responsible Computing}} \bibinfo{volume}{1}, \bibinfo{number}{2} (\bibinfo{year}{2024}), \bibinfo{pages}{11:1--11:52}.
\newblock


\bibitem[Hort et~al\mbox{.}(2021)]%
        {sigsoftHortZSH21}
\bibfield{author}{\bibinfo{person}{Max Hort}, \bibinfo{person}{Jie~M. Zhang}, \bibinfo{person}{Federica Sarro}, {and} \bibinfo{person}{Mark Harman}.} \bibinfo{year}{2021}\natexlab{}.
\newblock \showarticletitle{Fairea: A model behaviour mutation approach to benchmarking bias mitigation methods}. In \bibinfo{booktitle}{\emph{Proceedings of the 29th {ACM} Joint European Software Engineering Conference and Symposium on the Foundations of Software Engineering, ESEC/FSE 2021}}. \bibinfo{pages}{994--1006}.
\newblock


\bibitem[Kamiran and Calders(2011)]%
        {rewpaper}
\bibfield{author}{\bibinfo{person}{Faisal Kamiran} {and} \bibinfo{person}{Toon Calders}.} \bibinfo{year}{2011}\natexlab{}.
\newblock \showarticletitle{Data preprocessing techniques for classification without discrimination}.
\newblock \bibinfo{journal}{\emph{Knowledge and Information Systems}} \bibinfo{volume}{33}, \bibinfo{number}{1} (\bibinfo{year}{2011}), \bibinfo{pages}{1--33}.
\newblock


\bibitem[Kitchenham et~al\mbox{.}(2017)]%
        {eseKitchenhamMBKBC17}
\bibfield{author}{\bibinfo{person}{Barbara~A. Kitchenham}, \bibinfo{person}{Lech Madeyski}, \bibinfo{person}{David Budgen}, \bibinfo{person}{Jacky Keung}, \bibinfo{person}{Pearl Brereton}, \bibinfo{person}{Stuart~M. Charters}, \bibinfo{person}{Shirley Gibbs}, {and} \bibinfo{person}{Amnart Pohthong}.} \bibinfo{year}{2017}\natexlab{}.
\newblock \showarticletitle{Robust statistical methods for empirical software engineering}.
\newblock \bibinfo{journal}{\emph{Empirical Software Engineering}} \bibinfo{volume}{22}, \bibinfo{number}{2} (\bibinfo{year}{2017}), \bibinfo{pages}{579--630}.
\newblock


\bibitem[Li et~al\mbox{.}(2022)]%
        {icseLiMC0WZX22}
\bibfield{author}{\bibinfo{person}{Yanhui Li}, \bibinfo{person}{Linghan Meng}, \bibinfo{person}{Lin Chen}, \bibinfo{person}{Li Yu}, \bibinfo{person}{Di Wu}, \bibinfo{person}{Yuming Zhou}, {and} \bibinfo{person}{Baowen Xu}.} \bibinfo{year}{2022}\natexlab{}.
\newblock \showarticletitle{Training data debugging for the fairness of machine learning software}. In \bibinfo{booktitle}{\emph{Proceedings of the 44th {IEEE/ACM} International Conference on Software Engineering, {ICSE} 2022}}. \bibinfo{pages}{2215--2227}.
\newblock


\bibitem[Maheshwari et~al\mbox{.}(2023)]%
        {emnlp4BDK23}
\bibfield{author}{\bibinfo{person}{Gaurav Maheshwari}, \bibinfo{person}{Aur{\'{e}}lien Bellet}, \bibinfo{person}{Pascal Denis}, {and} \bibinfo{person}{Mikaela Keller}.} \bibinfo{year}{2023}\natexlab{}.
\newblock \showarticletitle{Fair without leveling down: {A} new intersectional fairness definition}. In \bibinfo{booktitle}{\emph{Proceedings of the 2023 Conference on Empirical Methods in Natural Language Processing, {EMNLP} 2023}}. \bibinfo{pages}{9018--9032}.
\newblock


\bibitem[McKnight and Najab(2010)]%
        {mcknight2010mann}
\bibfield{author}{\bibinfo{person}{Patrick~E McKnight} {and} \bibinfo{person}{Julius Najab}.} \bibinfo{year}{2010}\natexlab{}.
\newblock \showarticletitle{Mann-Whitney U test}.
\newblock \bibinfo{journal}{\emph{The Corsini Encyclopedia of Psychology}} (\bibinfo{year}{2010}), \bibinfo{pages}{1--1}.
\newblock


\bibitem[Menon and Williamson(2018)]%
        {fatMenonW18}
\bibfield{author}{\bibinfo{person}{Aditya~Krishna Menon} {and} \bibinfo{person}{Robert~C. Williamson}.} \bibinfo{year}{2018}\natexlab{}.
\newblock \showarticletitle{The cost of fairness in binary classification}. In \bibinfo{booktitle}{\emph{Proceedings of the Conference on Fairness, Accountability and Transparency, {FAT} 2018}}. \bibinfo{pages}{107--118}.
\newblock


\bibitem[Mittelstadt et~al\mbox{.}(2023)]%
        {mittelstadt2023unfairness}
\bibfield{author}{\bibinfo{person}{Brent Mittelstadt}, \bibinfo{person}{Sandra Wachter}, {and} \bibinfo{person}{Chris Russell}.} \bibinfo{year}{2023}\natexlab{}.
\newblock \showarticletitle{The unfairness of fair machine learning: Levelling down and strict egalitarianism by default}.
\newblock \bibinfo{journal}{\emph{arXiv preprint arXiv:2302.02404}} (\bibinfo{year}{2023}).
\newblock


\bibitem[Myers et~al\mbox{.}(2013)]%
        {myers2013research}
\bibfield{author}{\bibinfo{person}{Jerome~L Myers}, \bibinfo{person}{Arnold~D Well}, {and} \bibinfo{person}{Robert~F Lorch~Jr}.} \bibinfo{year}{2013}\natexlab{}.
\newblock \bibinfo{booktitle}{\emph{Research design and statistical analysis}}.
\newblock \bibinfo{publisher}{Routledge}.
\newblock


\bibitem[Norton and Sommers(2011)]%
        {norton2011whites}
\bibfield{author}{\bibinfo{person}{Michael~I Norton} {and} \bibinfo{person}{Samuel~R Sommers}.} \bibinfo{year}{2011}\natexlab{}.
\newblock \showarticletitle{Whites see racism as a zero-sum game that they are now losing}.
\newblock \bibinfo{journal}{\emph{Perspectives on Psychological science}} \bibinfo{volume}{6}, \bibinfo{number}{3} (\bibinfo{year}{2011}), \bibinfo{pages}{215--218}.
\newblock


\bibitem[Peng et~al\mbox{.}(2023)]%
        {fairmaskpaper}
\bibfield{author}{\bibinfo{person}{Kewen Peng}, \bibinfo{person}{Joymallya Chakraborty}, {and} \bibinfo{person}{Tim Menzies}.} \bibinfo{year}{2023}\natexlab{}.
\newblock \showarticletitle{FairMask: Better fairness via model-based rebalancing of protected attributes}.
\newblock \bibinfo{journal}{\emph{{IEEE} Transactions on Software Engineering}} \bibinfo{volume}{49}, \bibinfo{number}{4} (\bibinfo{year}{2023}), \bibinfo{pages}{2426--2439}.
\newblock


\bibitem[Pham et~al\mbox{.}(2020)]%
        {PhamQWLRTYN20}
\bibfield{author}{\bibinfo{person}{Hung~Viet Pham}, \bibinfo{person}{Shangshu Qian}, \bibinfo{person}{Jiannan Wang}, \bibinfo{person}{Thibaud Lutellier}, \bibinfo{person}{Jonathan Rosenthal}, \bibinfo{person}{Lin Tan}, \bibinfo{person}{Yaoliang Yu}, {and} \bibinfo{person}{Nachiappan Nagappan}.} \bibinfo{year}{2020}\natexlab{}.
\newblock \showarticletitle{Problems and opportunities in training deep learning software systems: an analysis of variance}. In \bibinfo{booktitle}{\emph{Proceedings of the 35th {IEEE/ACM} International Conference on Automated Software Engineering, {ASE} 2020}}. \bibinfo{pages}{771--783}.
\newblock


\bibitem[Tizpaz{-}Niari et~al\mbox{.}(2022)]%
        {icseNiariKT022}
\bibfield{author}{\bibinfo{person}{Saeid Tizpaz{-}Niari}, \bibinfo{person}{Ashish Kumar}, \bibinfo{person}{Gang Tan}, {and} \bibinfo{person}{Ashutosh Trivedi}.} \bibinfo{year}{2022}\natexlab{}.
\newblock \showarticletitle{Fairness-aware configuration of machine learning libraries}. In \bibinfo{booktitle}{\emph{Proceedings of the 44th {IEEE/ACM} International Conference on Software Engineering, {ICSE} 2022}}. \bibinfo{pages}{909--920}.
\newblock


\bibitem[Wolff(2001)]%
        {wolff2001levelling}
\bibfield{author}{\bibinfo{person}{Jonathan Wolff}.} \bibinfo{year}{2001}\natexlab{}.
\newblock \showarticletitle{Levelling down}.
\newblock In \bibinfo{booktitle}{\emph{Challenges to Democracy: Ideas, Involvement and Institutions}}. \bibinfo{publisher}{Springer}, \bibinfo{pages}{18--32}.
\newblock


\bibitem[Xiao et~al\mbox{.}(2024)]%
        {xiao2024mirrorfair}
\bibfield{author}{\bibinfo{person}{Ying Xiao}, \bibinfo{person}{Jie~M Zhang}, \bibinfo{person}{Yepang Liu}, \bibinfo{person}{Mohammad~Reza Mousavi}, \bibinfo{person}{Sicen Liu}, {and} \bibinfo{person}{Dingyuan Xue}.} \bibinfo{year}{2024}\natexlab{}.
\newblock \showarticletitle{MirrorFair: Fixing fairness bugs in machine learning software via counterfactual predictions}.
\newblock \bibinfo{journal}{\emph{Proceedings of the ACM on Software Engineering}} \bibinfo{volume}{1}, \bibinfo{number}{FSE} (\bibinfo{year}{2024}), \bibinfo{pages}{2121--2143}.
\newblock


\bibitem[Yang et~al\mbox{.}(2024)]%
        {zeyujjj}
\bibfield{author}{\bibinfo{person}{Junjie Yang}, \bibinfo{person}{Jiajun Jiang}, \bibinfo{person}{Zeyu Sun}, {and} \bibinfo{person}{Junjie Chen}.} \bibinfo{year}{2024}\natexlab{}.
\newblock \showarticletitle{A large-scale empirical study on improving the fairness of image classification models}. In \bibinfo{booktitle}{\emph{Proceedings of the 33rd {ACM} {SIGSOFT} International Symposium on Software Testing and Analysis, {ISSTA} 2024}}. \bibinfo{pages}{210--222}.
\newblock


\bibitem[Zafar et~al\mbox{.}(2017)]%
        {ZafarVGG17}
\bibfield{author}{\bibinfo{person}{Muhammad~Bilal Zafar}, \bibinfo{person}{Isabel Valera}, \bibinfo{person}{Manuel Gomez{-}Rodriguez}, {and} \bibinfo{person}{Krishna~P. Gummadi}.} \bibinfo{year}{2017}\natexlab{}.
\newblock \showarticletitle{Fairness beyond disparate treatment {\&} disparate impact: Learning classification without disparate mistreatment}. In \bibinfo{booktitle}{\emph{Proceedings of the 26th International Conference on World Wide Web, {WWW} 2017}}. \bibinfo{pages}{1171--1180}.
\newblock


\bibitem[Zhang et~al\mbox{.}(2018)]%
        {ADVpaper}
\bibfield{author}{\bibinfo{person}{Brian~Hu Zhang}, \bibinfo{person}{Blake Lemoine}, {and} \bibinfo{person}{Margaret Mitchell}.} \bibinfo{year}{2018}\natexlab{}.
\newblock \showarticletitle{Mitigating unwanted biases with adversarial learning}. In \bibinfo{booktitle}{\emph{Proceedings of the 2018 {AAAI/ACM} Conference on AI, Ethics, and Society, {AIES} 2018}}. \bibinfo{pages}{335--340}.
\newblock


\bibitem[Zhang and Harman(2021)]%
        {icseZhangH21}
\bibfield{author}{\bibinfo{person}{Jie~M. Zhang} {and} \bibinfo{person}{Mark Harman}.} \bibinfo{year}{2021}\natexlab{}.
\newblock \showarticletitle{Ignorance and prejudice in software fairness}. In \bibinfo{booktitle}{\emph{Proceedings of the 43rd {IEEE/ACM} International Conference on Software Engineering, {ICSE} 2021}}. \bibinfo{pages}{1436--1447}.
\newblock


\bibitem[Zhang et~al\mbox{.}(2022)]%
        {jieMLsurvey}
\bibfield{author}{\bibinfo{person}{Jie~M. Zhang}, \bibinfo{person}{Mark Harman}, \bibinfo{person}{Lei Ma}, {and} \bibinfo{person}{Yang Liu}.} \bibinfo{year}{2022}\natexlab{}.
\newblock \showarticletitle{Machine learning testing: Survey, landscapes and horizons}.
\newblock \bibinfo{journal}{\emph{{IEEE} Transactions on Software Engineering}} \bibinfo{volume}{48}, \bibinfo{number}{2} (\bibinfo{year}{2022}), \bibinfo{pages}{1--36}.
\newblock


\bibitem[Zhang and Sun(2022)]%
        {sigsoftZhang022}
\bibfield{author}{\bibinfo{person}{Mengdi Zhang} {and} \bibinfo{person}{Jun Sun}.} \bibinfo{year}{2022}\natexlab{}.
\newblock \showarticletitle{Adaptive fairness improvement based on causality analysis}. In \bibinfo{booktitle}{\emph{Proceedings of the 30th {ACM} Joint European Software Engineering Conference and Symposium on the Foundations of Software Engineering, {ESEC/FSE} 2022}}. \bibinfo{pages}{6--17}.
\newblock


\bibitem[Zheng et~al\mbox{.}(2022)]%
        {icseZhengCD0CJW0C22}
\bibfield{author}{\bibinfo{person}{Haibin Zheng}, \bibinfo{person}{Zhiqing Chen}, \bibinfo{person}{Tianyu Du}, \bibinfo{person}{Xuhong Zhang}, \bibinfo{person}{Yao Cheng}, \bibinfo{person}{Shouling Ji}, \bibinfo{person}{Jingyi Wang}, \bibinfo{person}{Yue Yu}, {and} \bibinfo{person}{Jinyin Chen}.} \bibinfo{year}{2022}\natexlab{}.
\newblock \showarticletitle{NeuronFair: Interpretable white-box fairness testing through biased neuron identification}. In \bibinfo{booktitle}{\emph{Proceedings of the 44th {IEEE/ACM} International Conference on Software Engineering, {ICSE} 2022}}. \bibinfo{pages}{1519--1531}.
\newblock


\bibitem[Zietlow et~al\mbox{.}(2022)]%
        {ZietlowLBKLS022}
\bibfield{author}{\bibinfo{person}{Dominik Zietlow}, \bibinfo{person}{Michael Lohaus}, \bibinfo{person}{Guha Balakrishnan}, \bibinfo{person}{Matth{\"{a}}us Kleindessner}, \bibinfo{person}{Francesco Locatello}, \bibinfo{person}{Bernhard Sch{\"{o}}lkopf}, {and} \bibinfo{person}{Chris Russell}.} \bibinfo{year}{2022}\natexlab{}.
\newblock \showarticletitle{Leveling down in computer vision: Pareto inefficiencies in fair deep classifiers}. In \bibinfo{booktitle}{\emph{Proceedings of the {IEEE/CVF} Conference on Computer Vision and Pattern Recognition, {CVPR} 2022}}. \bibinfo{pages}{10400--10411}.
\newblock


\end{thebibliography}

\end{document}